\documentclass[lettersize,journal]{IEEEtran}
\usepackage{amsmath,amsfonts}
\usepackage{algorithmic}
\usepackage{algorithm}
\usepackage{array}
\usepackage[caption=false,font=normalsize,labelfont=sf,textfont=sf]{subfig}
\usepackage{textcomp}
\usepackage{stfloats}
\usepackage{url}
\usepackage{verbatim}
\usepackage{graphicx}
\usepackage{cite}
\usepackage{multirow}

\usepackage{enumitem}
\usepackage{diagbox}
\usepackage{listings}
\usepackage{soul}
\usepackage{tabularx}
\usepackage{booktabs}
\usepackage{xcolor}
\makeatletter
\g@addto@macro\normalsize{%
  \setlength{\abovedisplayskip}{4pt plus 2pt minus 2pt}
  \setlength{\belowdisplayskip}{4pt plus 2pt minus 2pt}
  \setlength{\abovedisplayshortskip}{2pt plus 2pt}
  \setlength{\belowdisplayshortskip}{2pt plus 2pt minus 2pt}
}
\makeatother
\begin{document}
\setlength{\textfloatsep}{5pt plus 2pt minus 2pt} 
\setlength{\intextsep}{5pt plus 2pt minus 2pt}  
\setlength{\floatsep}{4pt plus 2pt minus 2pt}

\title{High-Fidelity 3D Facial Avatar Synthesis with Controllable Fine-Grained Expressions}

\author{Yikang~He,
        Jichao~Zhang,
        Wei~Wang,~\IEEEmembership{Member,~IEEE},
        Nicu~Sebe,~\IEEEmembership{Senior~Member,~IEEE},
        and Yao Zhao,~\IEEEmembership{Fellow,~IEEE},
\IEEEcompsocitemizethanks{\IEEEcompsocthanksitem Yikang He, Wei Wang, Yao Zhao are with the Institute of Information Science, Beijing Jiaotong University, Beijing, China. E-mail: 23120294@bjtu.edu.cn, wei.wang@bjtu.edu.cn, yzhao@bjtu.edu.cn.
\IEEEcompsocthanksitem Nicu Sebe is with the Department of Information Engineering and Computer Science (DISI), University of Trento, Italy. E-mail: sebe@disi.unitn.it.
\IEEEcompsocthanksitem Jichao Zhang is with the School of Computer Science, Ocean University of China, Qingdao, China. E-mail: zhang163220@gmail.com.}
}
\IEEEpubid{
  \makebox[\textwidth][c]{
    \begin{minipage}{\textwidth}
      \centering \scriptsize
      Copyright~\copyright~2026~IEEE. Personal use of this material is permitted. However, permission to use this material for any other purposes must be obtained from \\
      the IEEE by sending an email to pubs-permissions@ieee.org. 
      DOI: 10.1109/TCSVT.2026.3674337
    \end{minipage}
  }
}
\maketitle
\begin{abstract}
 Facial expression editing methods can be mainly categorized into two types based on their architectures: 2D-based and 3D-based methods. The former lacks 3D face modeling capabilities, making it difficult to edit 3D factors effectively. The latter has demonstrated superior performance in generating high-quality and view-consistent renderings using single-view 2D face images. Although these methods have successfully used animatable models to control facial expressions, they still have limitations in achieving precise control over fine-grained expressions. To address this issue, in this paper, we propose a novel approach by simultaneously refining both the latent code of a pretrained 3D-Aware GAN model for texture editing and the expression code of the driven 3DMM model for mesh editing. Specifically, we introduce a Dual Mappers module, comprising Texture Mapper and Emotion Mapper, to learn the transformations of the given latent code for textures and the expression code for meshes, respectively. To optimize the Dual Mappers, we propose a Text-Guided Optimization method, leveraging a CLIP-based objective function with expression text prompts as targets, while integrating a SubSpace Projection mechanism to project the text embedding to the expression subspace such that we can have more precise control over fine-grained expressions. Extensive experiments and comparative analyses demonstrate the effectiveness and superiority of our proposed method.
\end{abstract}


\section{Introduction}
\enlargethispage{-0.3in}
\IEEEPARstart{F}{acial} Expression Editing, aimed at manipulating the expression of a given face image to match a target expression without altering its identity properties, is increasingly prevalent across applications, such as object animation, human-computer interactions, psychological analysis, and augmented reality~\cite{siarohin2023unsupervised,li2023object,kosch2023survey,borsboom2021network,dargan2023augmented}. 

Generative models, including generative adversarial networks (GANs)~\cite{goodfellow2014generative} and diffusion models~\cite{song2020denoising}, have revolutionized facial attribute editing by enabling high-quality and flexible image synthesis. 

To address this task, early methods~\cite{pumarola2018ganimation,choi2018stargan,wu2020cascade,song2018geometry,he2019attgan,ding2018exprgan,abdal2021styleflow,patashnik2021styleclip,hu20232cet,huang2024interactive,10620351,10178013,10542240,zhang2024open} predominantly relied on 2D image translation techniques. However, these approaches often fail to account for 3D facial geometry, limiting their ability to maintain 3D consistency and control pose variations. Consequently, recent advancements have integrated 3D Morphable Face Models (3DMM)~\cite{blanz1999morphable} with GANs or Neural Radiance Fields (NeRF) to enable finer control.

Despite these improvements, achieving high-fidelity editing with both view consistency and fine-grained expression control remains an open problem. 3D GAN methods, such as AniFaceGAN~\cite{wu2022anifacegan} and Next3D~\cite{sun2023next3d}, combine volume rendering with mesh-guided deformation. However, they heavily rely on 3D reconstruction priors like DECA~\cite{DECA:Siggraph2021}, which lack rich fine-grained expression variations in their training data, limiting the model's ability to extract and generate subtle expression details.

Furthermore, while diffusion-based methods like DiffusionRig~\cite{ding2023diffusionrig} excel in image fidelity, they lack explicit 3D representations, often resulting in compromised multi-view consistency compared to geometry-aware approaches. Although multi-view diffusion methods like Morphable Diffusion~\cite{chen2024morphable}, incorporating 3D morphable models, can mitigate inconsistency, they typically require extensive multi-view datasets for each individual, restricting their practicality in data-scarce scenarios. Similarly, recent reconstruction-based approaches, such as Gaussian Head Avatars~\cite{xu2024gaussian} or methods utilizing SDS optimization~\cite{10963753}, necessitate training separate models for each individual (per-subject optimization). This requirement makes them computationally inefficient and time-consuming for general applications.
\enlargethispage{-0.3in}
In contrast, to address the aforementioned gaps, the primary research problem we tackle in this paper is achieving high-fidelity, fine-grained 3D Facial expression editing from a single image while strictly maintaining multi-view consistency. By leveraging a robust 3D GAN-based framework, we eliminate the need for multi-view data or per-subject training. This approach efficiently ensures view consistency while overcoming the common challenge of uncoordinated texture and geometry updates—where existing methods fail to synergistically match structural mesh deformations with subtle texture changes, often leading to unnatural editing artifacts.

\begin{figure*}
\centering
  \includegraphics[width=0.9\textwidth]{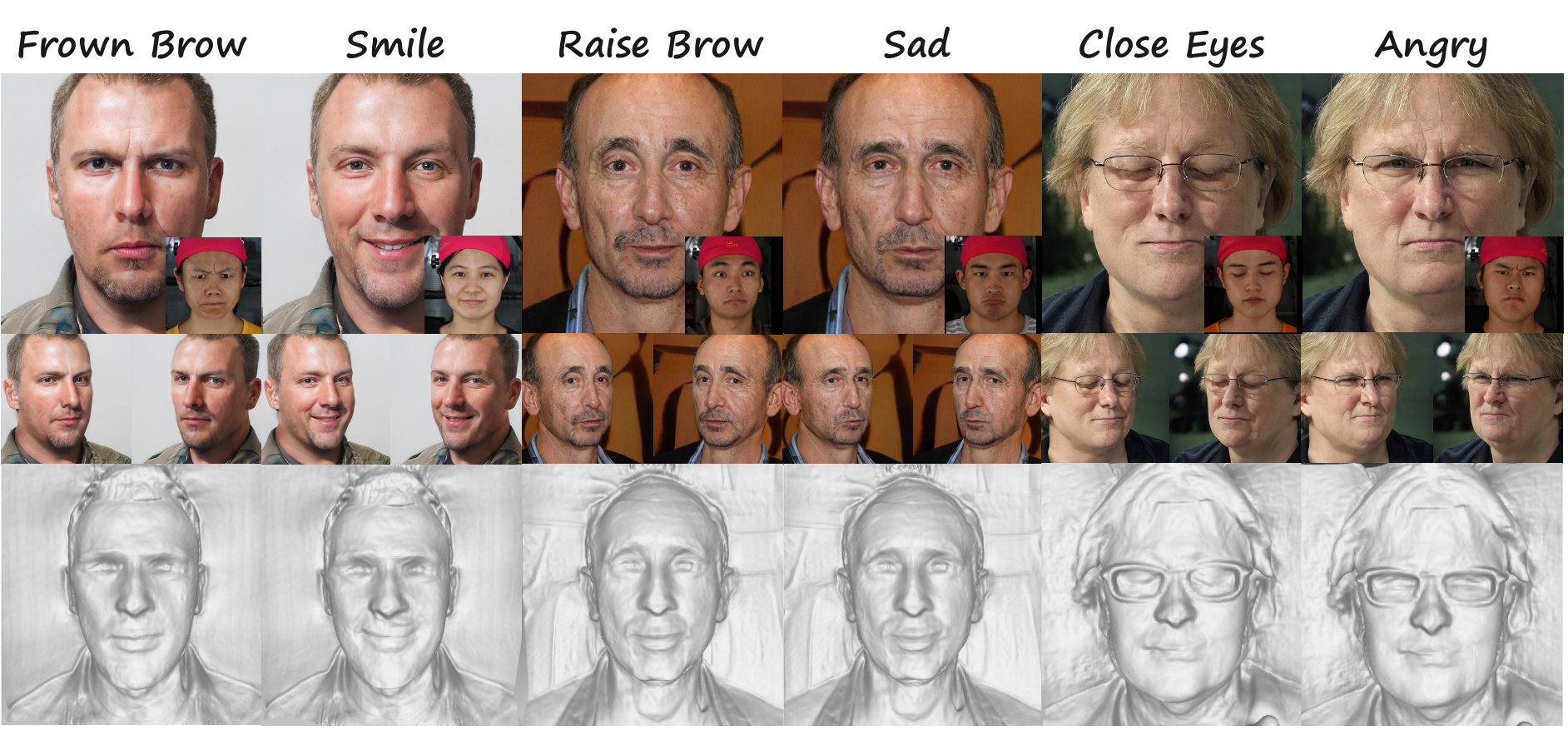}
  \caption{Our model can generate high-quality, view-consistent face images while enabling fine-grained expression editing. The first row displays the generated images of different facial expression edits (with the reference image located in the lower-right corner), the second row presents different viewpoints, and the third row showcases the generated 3D head meshes.}
  \label{fig:teaser}
  \vspace{-12pt}
\end{figure*}

Fine-grained facial expressions, capturing subtle emotional nuances, present challenges in both recognition and generation due to the lack of comprehensive annotations. This scarcity hinders expression editing and generation tasks. To address this, we leverage CLIP~\cite{radford2021learning}, a multimodal model trained on image-text pairs. CLIP's ability to generalize across visual and linguistic domains allows it to guide fine-grained expression manipulation through multimodal supervision. By using expression descriptions (\emph{e.g.}, ``A person who is raising brow'') as supervision via CLIP loss, we can bypass the need for large annotated datasets, enabling the generation of high-quality 3D avatars with precise expressions.

In this paper, to solve the fine-grained expression editing problem, we introduce a novel 3D-aware framework designed to synergistically edit both texture and geometry. 

First, given a reference image, we use EMOCA~\cite{EMOCA:CVPR:2021}, which has a richer expression space, to extract its 3D shape hyper-parameters (\emph{i.e.} FLAME parameters) for initialization. Next, to ensure this coordinated editing, we introduce a \textbf{Dual Mappers Module}, comprising a Texture Mapper for refining the UV texture latent code $\boldsymbol{w}$ and an Emotion Mapper for optimizing the expression code  $\boldsymbol{\alpha}$ of the mesh. Both components work simultaneously and interact via a cross-attention mechanism. By synchronously updating both spaces, our method ensures that geometric and textural features match coherently, significantly reducing artifacts introduced during the editing process and achieving more accurate expression synthesis.

To train the Dual Mappers, we propose a \textbf{Text-Guided Optimization method} that utilizes CLIP-based loss functions, enabling expression refinement based on textual prompts. 
Note that each textual prompt corresponds to one expression. For each individual expression, we learn one set of specific parameters for Dual Mappers.
This method reduces reliance on extensively annotated fine-grained expression datasets by leveraging the generalization capabilities of large pre-trained language-vision models. It effectively bridges the gap between textual descriptions and visual representations, providing a robust mechanism for controlling subtle expression nuances. Furthermore, a Subspace Projection mechanism ensures the preservation of identity attributes, maintaining the fidelity of the synthesized face while allowing fine-grained expression adjustments. As shown in Fig.~\ref{fig:teaser}, guided by reference images and target expression texts, our method is capable of generating high-quality multi-view facial images while effectively capturing both the texture and geometric characteristics of the target expression.

By addressing these limitations and leveraging the proposed techniques, our approach advances the field of 3D-aware expression editing and opens new possibilities for realistic and expressive 3D facial synthesis in practical applications.

To sum up, the main contributions are as follows:
\begin{itemize}[]
    \item We have proposed a novel Dual Mappers module for face synthesis with fine-grained expression editing which does not require large scale training data.
    \item The Dual Mappers comprise a Texture Mapper, which refines the latent code of UV textures, and an Emotion Mapper, which learns transformations of the expression parameters of the mesh. Besides, a cross-attention mechanism has been designed for their interactions, and more accurate expression editing can be achieved.
    \item We have proposed a Text-Guided Optimization for Dual Mappers, utilizing the CLIP-based Loss with the text prompts as the objective function, while integrating a Subspace Projection mechanism to avoid undesired attribute changes. 
\end{itemize}

\section{RELATED WORKS}
\noindent\textbf{3D-Aware Head-Avatar Generation.}
Recent advancements in 3D scene representation methods have improved 3D-aware image synthesis for head avatar generation. Many models are trained on 2D images without relying on 3D or multiview data, using techniques like Explicit Voxel Grids~\cite{gadelha20173d, henzler2019escaping}, Neural Implicit Representations (NeRF)~\cite{schwarz2020graf, niemeyer2021giraffe,zhu2024dfie3d} or 3D Gaussian Splatting (3DGS)~\cite{hu2025tgavatar,xu2024gaussian,zhang2025humanref}. NeRF-based GANs allow for higher-resolution modeling without visible artifacts. Studies have introduced improvements in articulated object modeling~\cite{zhang20223d}, fast rendering~\cite{schwarz2022voxgraf}, and controllability for editing tasks~\cite{wu2022anifacegan}. 

For 3D avatar generation, leveraging a 3D morphable face model~\cite{galanakis20233dmm,li2017learning} within a 3D generative model enhances the controllability of pose and expression in models~\cite{tang2022explicitly,wu2022anifacegan,deng2020disentangled,sun2023next3d,yu2024ecavatar,10947104,10236465,9999364,10510337,li2025vividlistener}. Specifically, DiscoFaceGAN~\cite{deng2020disentangled} uses imitative-contrastive learning and the 3D Morphable Model (3DMM) for precise attribute control but struggles with pose-inconsistent issues. AniFaceGAN~\cite{wu2022anifacegan} uses expression-driven deformation fields for realistic avatar generation but faces challenges with fine-grained animation, such as gaze, and low-quality generation in regions like the mouth. These issues arise from the under-constrained nature of the 3DMM and deformation fields. Next3D~\cite{sun2023next3d} overcomes these challenges by separating dynamic and static components of the head and modeling them separately, using neural textures for facial deformation and an efficient teeth synthesis module. However, it struggles with expression control due to the limitations of the FLAME~\cite{FLAME:SiggraphAsia2017} and DECA~\cite{DECA:Siggraph2021} estimation methods.

\noindent\textbf{Facial Expression Editing.}
Several studies have achieved notable facial expression editing using generative adversarial networks (GANs), broadly categorized into 2D-GAN methods~\cite{pumarola2018ganimation,choi2018stargan,9625034,9915620,sun2024anyface++}, 3DMM-based methods~\cite{deng2020disentangled,geng20193d}, and NeRF-based methods~\cite{sun2022fenerf,jiang2022nerffaceediting}. 2D-GAN methods incorporate an encoder to map source to target domains for facial attribute manipulation, but they are limited by low-resolution outputs. StyleFlow~\cite{abdal2021styleflow} and StyleCLIP~\cite{patashnik2021styleclip} enhance visual quality by manipulating pretrained StyleGAN's latent space but lack 3D understanding. 3DMM-based methods~\cite{deng2020disentangled} use 3D Morphable Model (3DMM) priors for editing, though they struggle with pose control and view-inconsistency. More recently, NeRF-based methods~\cite{sun2022fenerf,jiang2022nerffaceediting} use semantic masks or 3DMM priors for facial expression control, but fine-grained expression control remains difficult due to limitations in capturing subtle variations in NeRF representations.

Recently, diffusion models~\cite{ho2020denoising,song2020denoising} have demonstrated superior performance compared to GANs, significantly enhancing image fidelity across diverse tasks. DiffusionRig~\cite{ding2023diffusionrig} introduces a two-stage training method, utilizing diffusion to learn personalized facial priors on top of generic face priors, enabling the facial expression editing task. However, due to lacking multiple-view data for training, the utilizing diffusion model fails to preserve the view-consistency. Morphable Diffusion~\cite{chen2024morphable} introduces a novel method for avatar creation by integrating a 3D morphable model into a multi-view consistent diffusion framework. However, it heavily relies on large amounts of multi-view image training data, making it difficult to generalize to images outside of the dataset.

\noindent\textbf{3D Face Reconstruction and 3DMM.} The 3D Morphable Model (3DMM) was first introduced by Blanz and Vetter~\cite{blanz1999morphable}, using PCA to model 3D face shapes and textures from scan data. Paysan et al.\cite{paysan20093d} further refined this with the Basel Face Model (BFM), employing advanced scanning technology for higher accuracy. In 2017, Gerig et al.\cite{gerig2018morphable} enhanced BFM by adding facial expression modeling. Other advancements included multilinear and non-linear representations for 3DMM~\cite{li2020learning, li2010example, neumann2013sparse, brunton2014multilinear, tran2019towards, tran2018nonlinear}, while Li et al.~\cite{li2017learning} introduced the FLAME model, which accurately simulates head poses and eye rotations.

Reconstructing 3D facial shapes from images has long been a challenge. One approach is regressing 3DMM parameters to generate 3D faces~\cite{deng2020retinaface, deng2019accurate, DECA:Siggraph2021, EMOCA:CVPR:2021, wei20193d, deng2022end,deng2022end,9832921}, which allows controlled editing of facial meshes. However, it’s difficult to obtain large labeled datasets for 3D meshes, making self-supervised methods more appealing. Deep3D~\cite{deng2019accurate} regresses BFM parameters for precise 3D face reconstruction from multi-view images, but it lacks full head and neck modeling. To address this, DECA adopted the FLAME model~\cite{li2017learning} for improved animation and detail, using a two-stage training process to output animated, detailed shapes from a single image. However, DECA’s loss functions were insufficient for capturing facial expressions. EMOCA~\cite{EMOCA:CVPR:2021, filntisis2022visual} solved this by employing an expression recognition network and a dedicated encoder for more accurate facial expressions. Building upon this, we refined expression codes using CLIP~\cite{radford2021learning} for fine-grained control over facial expressions.
\begin{figure*}[t]
\centering
    \includegraphics[width=0.95\textwidth]{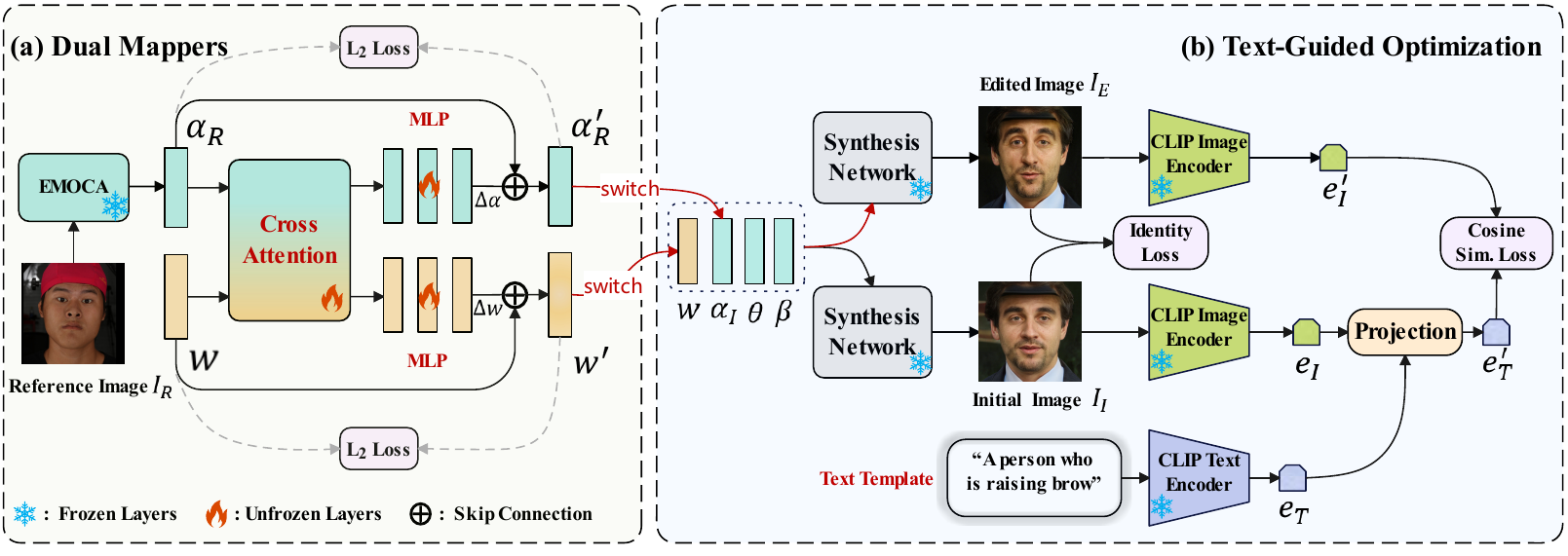}
    \caption{Overview of the proposed architecture which consists of two main modules: \textbf{(a) Dual Mappers} and \textbf{(b) Text-Guided Optimization}. 
    In (a), \textit{Dual Mappers} refine the random latent code $\boldsymbol{w}$ and expression code $\boldsymbol{\alpha_{R}}$ from a reference image $\boldsymbol{I_{R}}$ using a Cross Attention module and MLP layers, with updates stabilized by skip connections and a \textbf{$L_2$ Loss}. The refined codes $\boldsymbol{w^\prime}$ and $\boldsymbol{\alpha_{R}^\prime}$ are passed to a frozen synthesis network to generate the edited image. (The red arrow represents replacing $\boldsymbol{w}$,$\boldsymbol{\alpha_{I}}$ with $\boldsymbol{w^\prime}$,$\boldsymbol{\alpha_{R}^\prime}$.)
    In (b), \textit{Text-Guided Optimization} aligns the edited image with a text description (\emph{e.g.}, ``A person who is raising brow''). A \textbf{Projection} module maps the text embedding $\boldsymbol{e_{T}}$ to the image embedding space, enhancing compatibility. \textbf{Cosine Similarity Loss} aligns the embeddings, while \textbf{Identity Loss} preserves facial identity.}

    \label{figure:network}
    \vspace{-10pt}
  \end{figure*}

\begin{figure}[t]
\centering
    \includegraphics[width=0.95\linewidth]{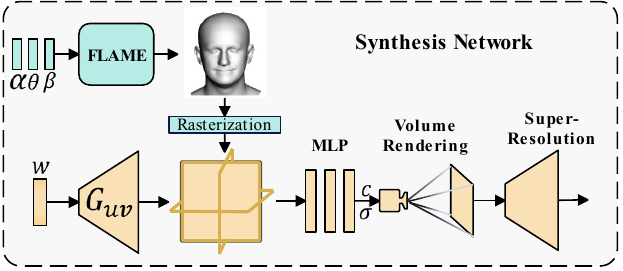}
    \caption{The \textit{Synthesis Network} utilizes Tri-Plane and Volume Rendering to achieve 3D-aware head avatar generation. The generator $\boldsymbol{G_{uv}}$ produces a Tri-Plane from the latent code $\boldsymbol{w}$, while FLAME accepts $\boldsymbol{\alpha}$,$\boldsymbol{\theta}$,$\boldsymbol{\beta}$ as input and provides a coarse head mesh aligned via rasterization. An MLP predicts color $\boldsymbol{c}$ and density $\boldsymbol{\sigma}$, which are used for \textbf{Volume Rendering}. Finally, a \textbf{Super-Resolution} module enhances the output, creating realistic, editable 3D avatars.}
    \label{figure:Synthesis}
  \end{figure}
\section{Method}

In this section, we present our novel framework for fine-grained, text-guided 3D facial expression editing. As illustrated in Fig.~\ref{figure:network}, our approach is built upon a 3D-aware generative backbone and consists of two core modules designed to bridge the gap between semantic text instructions and 3D geometry control: \textbf{(a) Dual Mappers} and \textbf{(b) Text-Guided Optimization}.

First, to achieve high-fidelity 3D face modeling with precise expression capture, we adopt Next3D~\cite{sun2023next3d} as our synthesis network but integrate EMOCA~\cite{EMOCA:CVPR:2021} to replace the standard coarse expression estimator. This modification allows for more accurate extraction of fine-grained expression parameters from reference images.
Second, we introduce the \textit{Dual Mappers} module designed to explicitly edit both the latent style code $\boldsymbol{w}$ and the expression parameters $\boldsymbol{\alpha}$. By employing a cross-attention mechanism, this module modulates these representations to ensure that both texture and geometric deformations are coherently altered to satisfy target expressions.
Finally, to train the Dual Mappers for accurate expression editing, we propose a \textit{Text-Guided Optimization} strategy. Specifically, to interpret textual descriptions (e.g., ``A person who is raising brow''), we utilize a subspace projection technique to disentangle expression-related attributes from identity features. This optimization objective guides the Dual Mappers to precisely modify facial expressions according to the text prompts while preserving the subject's identity.
\vspace{-10pt}
\subsection{Mesh-Guided 3D-aware Head Avatar}
\subsubsection{3D Face Modeling} 
In this work, we employ the FLAME~\cite{FLAME:SiggraphAsia2017} model as 3D prior for the generation and manipulation of various geometric shapes and fine-grained facial expressions. Taking a set of identity shape parameters $\boldsymbol{\theta}\in\mathbb{R}^{|\boldsymbol{\theta}|}$, pose parameters $\boldsymbol{\beta}\in\mathbb{R}^{|\boldsymbol{\beta}|}$, and facial expression parameters $\boldsymbol{\alpha}\in\mathbb{R}^{|\boldsymbol{\alpha}|}$ as input, FLAME outputs a corresponding mesh and is defined as:
\begin{equation}  M\left(\boldsymbol{\theta},\boldsymbol{\beta},\boldsymbol{\alpha}\right)\to\left(\mathbf{V},\mathbf{F}\right)
    \label{eq:flame}
  \end{equation}
with $n_v$ = 5023 vertices $\mathbf{V}\in\mathbb{R}^{n_{v}\times3}$ and $n_f$ = 9976 faces $\mathbf{F}\in\mathbb{R}^{n_{f}\times3}$. To reconstruct FLAME parameters from a single image, we employ pretrained EMOCA~\cite{EMOCA:CVPR:2021} as the parameter extractor instead of DECA~\cite{DECA:Siggraph2021} utilized in Next-3D~\cite{sun2023next3d}. Trained on Affectnet~\cite{mollahosseini2017affectnet}, a large-scale annotated emotion dataset, EMOCA incorporates an additional expression encoder to reconstruct expression parameters $\boldsymbol{\alpha}$.
This enhances its capability to capture fine-grained expressions and allows for more precise reconstruction of the required FLAME parameters. Given an image ${I}$, EMOCA outputs FLAME geometry parameters $\boldsymbol{\theta}$, $\boldsymbol{\beta}$, $\boldsymbol{\alpha}$, 
expressed as follows:
\begin{equation}
E(I)\to(\boldsymbol{\theta},\boldsymbol{\beta},\boldsymbol{\alpha})
\label{eq:coarse encoder}
\end{equation}
Using the above method enables the 3D face modeling from a single image while simultaneously allowing for precise representation of fine-grained emotions.

\subsubsection{Synthesis Network} 
As mentioned above, we adopt the pretrained Next3D~\cite{sun2023next3d} as the Synthesis Network, as shown in Fig. \ref{figure:Synthesis}. Next3D employs a StyleGAN2 CNN generator $G_{uv}$ to generate UV textures. These textures, combined with the 3D mesh obtained from the FLAME model through rasterization rendering, subsequently transform into neural texture tri-planes. Moreover, the final image is generated using volume rendering and a 2D super-resolution module. Specifically, it starts by sampling the latent code $\boldsymbol{z}\sim N(0,1)$ and then employs a Mapping Network to obtain an intermediate latent code $\boldsymbol{w}$ which is mapped into uv texture features $F_{uv}$ by $G_{uv}$:   
\begin{equation}
F_{uv}=G_{uv}(\boldsymbol{w})
\end{equation}
Then, Next3D uses DECA~\cite{DECA:Siggraph2021} to reconstruct the input image to obtain the 3D mesh $M$. However, as illustrated in Fig. \ref{figure:network}, we employ EMOCA~\cite{EMOCA:CVPR:2021} to perform this operation. Employing rasterization techniques $R$, the frontal, lateral, and top views of this mesh are rendered onto the generated UV texture maps. This leads to the generation of neural texture tri-planes:
\begin{equation}
F_{tri}=R(F_{uv},M(\boldsymbol{\theta},\boldsymbol{\beta},\boldsymbol{\alpha}))
\end{equation}
Next, by sampling features from the tri-planes $F_{tri}$, the density $\sigma$ and color $c$ can be obtained with a mapping network. To conserve computational resources, Next3D first utilizes volume rendering with these two parameters to generate a low-resolution image, which is then processed through a super-resolution module to generate the final RGB image.
\vspace{-12pt}

\subsection{Dual Mappers}
\label{Dual Mappers}
The latent space $\mathcal{W}$ of the pretrained StyleGAN2 generator serves as a rich representation that encapsulates diverse facial attributes and texture features. By modifying the latent code $\boldsymbol{w}$, it is possible to alter the identity, facial attributes, and texture features of the generated 3D head. Simultaneously, the facial expression parameters $\boldsymbol{\alpha}$ control the geometric configuration of the corresponding 3D mesh, enabling precise adjustments to facial expressions. Consequently, we devise a dual-mapping structure employing cross-attention mechanisms to predict transformations for the input $\boldsymbol{w}$ and $\boldsymbol{\alpha}$, thereby enabling fine-grained control over facial expression editing.
\begin{figure*}[t]
\centering
    \includegraphics[width=0.95\textwidth]{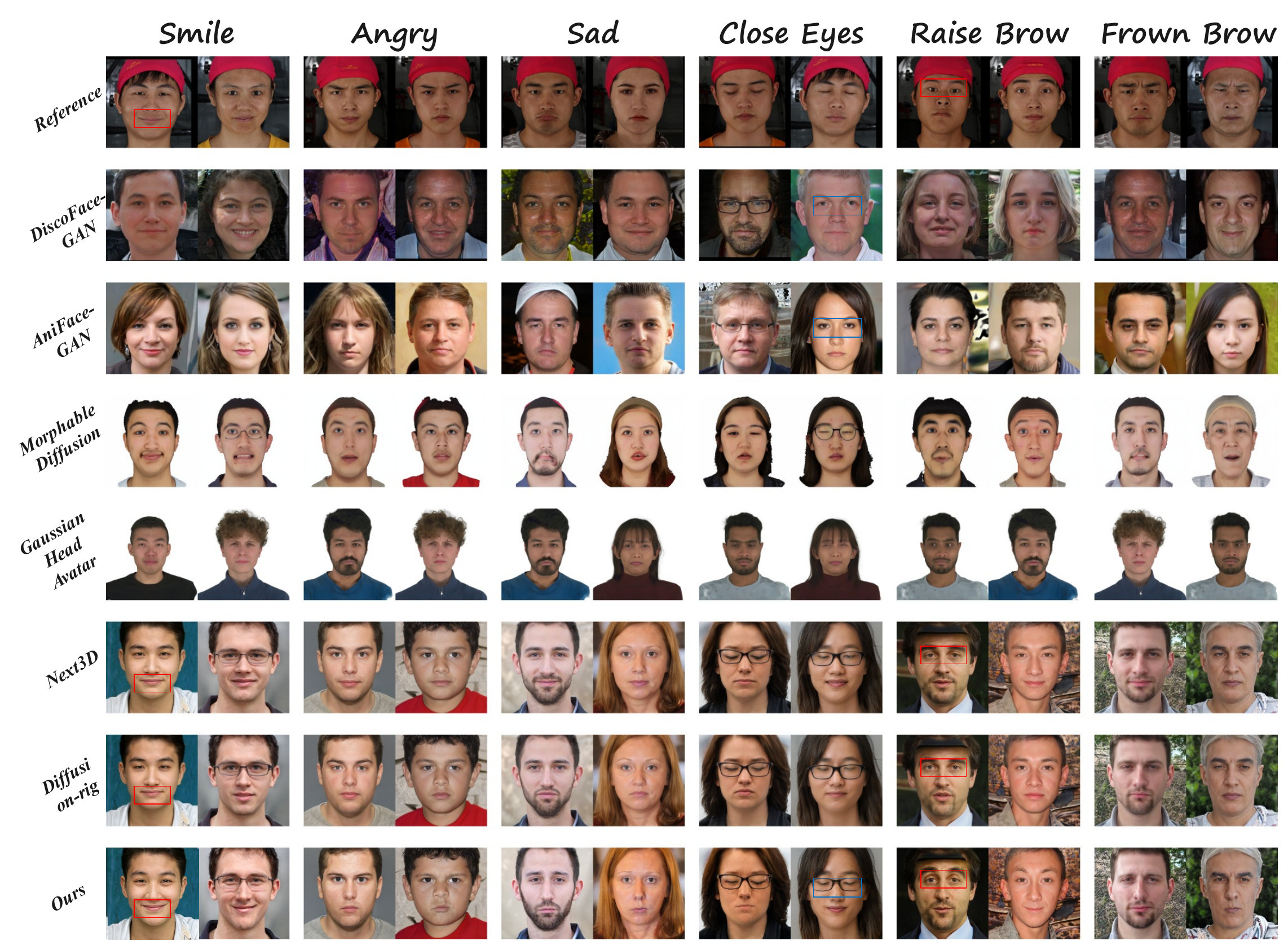}
    \caption{Fine-grained expression editing comparison with state-of-the-art animatable 3D image synthesis methods. To ensure a strict and direct comparison, methods supporting reference-guided editing within the same domain (Next3D, Diffusion-rig, and Ours) are evaluated using the exact same identity. For unconditional generative models (DiscoFaceGAN, AniFaceGAN) and models trained on entirely disparate datasets (Morphable Diffusion on FaceScape~\cite{yang2020facescape}, Gaussian Head Avatar on NeRSemble~\cite{kirschstein2023nersemble}), forcing a specific target identity would require error-prone inversion or cross-domain adaptation. To avoid disadvantaging these baselines with reconstruction artifacts, we display high-quality native identities sampled directly from their respective distributions. Fig.~\ref{figure:zoom-in} illustrates a zoomed comparison of the red-box region.}
    \vspace{-10pt}
    \label{figure:3D}
  \end{figure*}
As shown in Fig.~\ref{figure:network} (a), the Dual Mappers consist of a Texture Mapper $\mathcal{M}_T$ and an Emotion Mapper $\mathcal{M}_E$, along with a Cross Attention module. Owing to the aggregation of latent codes $\boldsymbol{w}$ and $\boldsymbol{\alpha}$ within the synthesis network through rasterization rendering on the neural texture tri-planes, we employ cross-attention to guide the editing of $\boldsymbol{\alpha}$ and $\boldsymbol{w}$ with $\boldsymbol{w}$ and $\boldsymbol{\alpha}$ as conditions, respectively. In addition, due to the significant impact that subtle numerical changes in the latent code can significantly influence the subsequent generation results, we refrain from directly predicting the edited parameters. Instead, we predict the discrepancy information between the edits pre- and post-editing. The Texture Mapper $\mathcal{M}_T$ is employed to predict $\Delta \boldsymbol{w}$, while the Emotion Mapper $\mathcal{M}_E$ predicts $\Delta \boldsymbol{\alpha}$. 

Specifically, as shown in Fig.~\ref{figure:network} (a), we first extract the facial expression parameters $\boldsymbol{\alpha}_{R}$ of the reference image $I_R$ using EMOCA~\cite{EMOCA:CVPR:2021}. Then, to derive the transformation $\Delta \boldsymbol{w}$, we utilize $\boldsymbol{\alpha}_R$ as the query and $\boldsymbol{w}$ as the key and value, resulting in the attention map $A_T$. Subsequently, $\Delta \boldsymbol{w}$ is obtained through mapping by $\mathcal{M}_T$, formulated as:
\begin{equation}
A_T=\mathrm{softmax}(\frac{Q_{\boldsymbol{\alpha}}K_{\boldsymbol{w}}^\top}{\sqrt{d}}),    \qquad                           \Delta\boldsymbol{w}=\mathcal{M}_T(A_TV_{\boldsymbol{w}}),
\label{eq:texture mapper}
\end{equation}
where $Q_{\boldsymbol{\alpha}}$ is mapped from $\boldsymbol{\alpha}_{R}$, $K_{\boldsymbol{w}}$ and $V_{\boldsymbol{w}}$ are mapped from $\boldsymbol{w}$, $d$ denotes the length of the key and query features. Then the target latent code $\boldsymbol{w}^\prime$ can be derived with $\Delta \boldsymbol{w} + \boldsymbol{w}$. By a similar approach but in reverse, utilizing $\boldsymbol{w}$ as the condition, we can derive $\Delta \boldsymbol{\alpha}$ and $\boldsymbol{\alpha}^\prime$ using $\mathcal{M}_E$.

Moreover, to prevent the edited parameters from deviating significantly from their initial values, thereby causing changes in other facial attributes or increasing artifacts in the generated images, we apply an $L_2$ loss term to constrain the Dual Mappers as follows:
\begin{equation}
\mathcal{L}_{\mathrm{M}}=||\boldsymbol{w}-\boldsymbol{w}^\prime||_2+||\boldsymbol{\alpha}- \boldsymbol{\alpha}^\prime||_2
\end{equation}
\vspace{-12pt}

\subsection{Text-Guided Optimization}
Recent studies~\cite{patashnik2021styleclip,kocasari2022stylemc,xia2021tedigan,aneja2023clipface} have increasingly showcased the efficacy of editing latent space $\mathcal{W}$ from StyleGAN under CLIP guidance, enabling high-fidelity image generation and precise control over editing directions. Therefore, we leverage the rich semantic information embedded in CLIP to guide the editing directions of the Dual Mappers outlined in Section~\ref{Dual Mappers} through textual cues. Additionally, some scholars~\cite{zhou2023clip} have noted that the text embedding obtained from CLIP may not always align well with the image embedding, possibly resulting in undesired attribute alterations during text-guided image editing and increased artifact presence in generated images. Inspired by their work, we abstain from directly optimizing our model through cosine similarity computation between the edited image embedding and the text description embedding. Instead, as shown in Fig.~\ref{figure:network} (b), guided by text embedding $\boldsymbol{e}_T$, we project the initial image embedding $\boldsymbol{e}_I$ onto a subspace $\mathfrak{T}$ composed of some basis vectors corresponding to a set of predefined text prompts. Then we calculate the cosine similarity between $\boldsymbol{e}_T^\prime$ which is obtained from the projection operation and the edited image embedding $\boldsymbol{e}_I^\prime$ to optimize our Dual Mappers.
\begin{figure}[t]

\centering
    \includegraphics[width=0.95\linewidth]{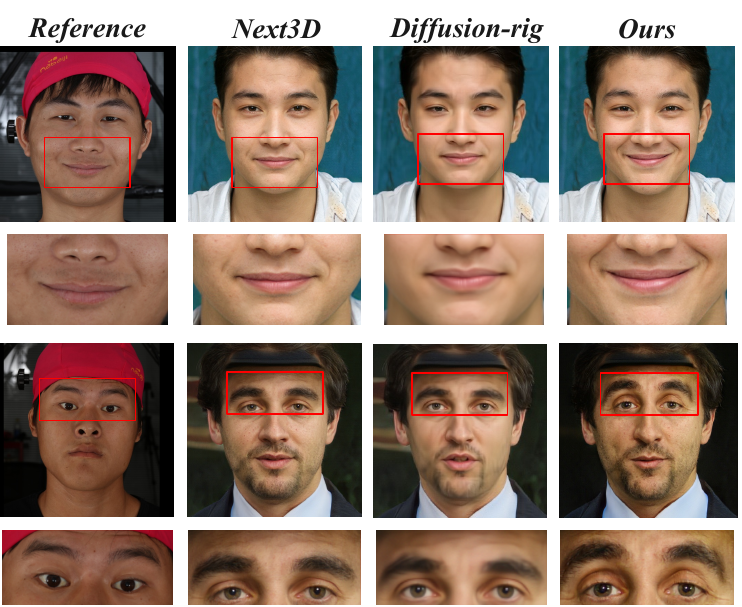}
    \caption{In contrast to Next3D and Diffusion-rig, our approach excels in capturing fine-grained expression facial details, such as the curvature of the mouth corners during smiling and the positioning of the eyebrows during frowning, bringing generated facial features closer to the reference image and target emotion text descriptions.
}
    \label{figure:zoom-in}
  \end{figure}
Specifically, since our primary target attribute for editing is facial expression, we choose some text descriptions (e.g., ``A person who is raising brow.'') related to expressions as basis vectors $\{\boldsymbol{b}_k\}_{k=1}^N$ to construct the subspace $\mathfrak{T}$. We then project the initial image embedding $\boldsymbol{e}_I$ into $\mathfrak{T}$ while preserving a residual vector $\boldsymbol{r}$, formulated as:
\begin{equation}
\boldsymbol{e}_P=\mathcal{P}_{\mathfrak{T}}(\boldsymbol{e}_I),\qquad   \boldsymbol{r}=\boldsymbol{e}_I-\boldsymbol{e}_P,
\label{eq:projection}
\end{equation}
where $\mathcal{P}$ is the projection operation on $\mathfrak{T}$, $\boldsymbol{e}_P$ is the projected embedding. After that, we augment the impact of the text prompt $t$ on the projected embedding $\boldsymbol{e}_P$, and subsequently, reintroduce the residual vector $\boldsymbol{r}$ to form the final embedding:
\begin{equation}
\boldsymbol{e}_T^\prime=\mathcal{A}(\boldsymbol{e}_P,\boldsymbol{e}_T,\gamma)+\boldsymbol{r},
\label{eq:augmentation}
\end{equation}
where $\mathcal{A}$ is the augmentation operation performed on $\boldsymbol{e}_P$ under the guidance of $\boldsymbol{e}_T$, $\gamma\in\mathbb{R}^+$ is the augmentation power. The principle is to weaken attributes in $\boldsymbol{e}_P$ that are unrelated to $\boldsymbol{e}_T$, thereby enhancing the influence of the attributes represented by $\boldsymbol{e}_T$. Specifically,  $\mathcal{A}$ is defined as:
\begin{equation}
    \mathcal{A}(\boldsymbol{e}_P,\boldsymbol{e}_T,\gamma):=\sum_{k=1}^N\left(c_k-\gamma\left|c_k\right|\right)\boldsymbol{b}_k+\frac{\gamma\sum_{k=1}^N\left|c_k\right|}{\sum_{k=1}^Nd_k}\boldsymbol{e}_T,
\end{equation}
where $c_k=\boldsymbol{e}_P^T\boldsymbol{b}_k$, $d_k=\boldsymbol{e}_T^T\boldsymbol{b}_k.$, and $\boldsymbol{b}_k$ is a set of basis vectors, each describing an expression.

In summary, after projecting $\boldsymbol{e}_I$ into the expression subspace $\mathfrak{T}$, text-guided facial editing focuses only on attributes related to expressions, ensuring that other attributes remain unchanged. By optimizing the cosine similarity between the final embedding $\boldsymbol{e}_T^\prime$ and the edited image embedding $\boldsymbol{e}_I^\prime$, fine-grained expression editing under text guidance can be achieved. Additionally, we introduce an identity loss to constrain identity changes during editing, formulated as:
\begin{equation}
\mathcal{L}_{\mathrm{CLIP}} = -\left<\boldsymbol{e}_I^\prime,\boldsymbol{e}_T^\prime\right> \quad
\mathcal{L}_{\mathrm{ID}} = 1-\left<FR(I_I),FR(I_E)\right>
\label{eq:cosine similarity}
\end{equation}
where $FR$ represents a pretrained ArcFace~\cite{deng2019arcface} network utilized for face recognition, and $\langle\cdot,\cdot\rangle $ calculates the cosine similarity between its arguments. The final optimization objective is given by:
\begin{equation}
\mathcal{L}_{\mathrm{Total}}=\lambda_{\mathrm{CLIP}}\mathcal{L}_{\mathrm{CLIP}}+\lambda_{\mathrm{M}}\mathcal{L}_{\mathrm{M}}+\lambda_{\mathrm{ID}}\mathcal{L}_{\mathrm{ID}}
\label{eq:total loss}
\end{equation}
where the hyperparameters controlling the relative weights of each loss term are empirically set to $\lambda_{\mathrm{CLIP}} = 1$, $\lambda_{\mathrm{M}} = 0.05$, and $\lambda_{\mathrm{ID}} = 0.2$.
\begin{table*}[t]
\setlength\tabcolsep{2pt}
\caption{Quantitative results on the expression editing results using two metrics: Action Units Accuracy (AU Acc) and CLIP Score. The used expressions are Smile, Angry, Sad, Close Eyes (CE), Raise Brow (RB), and Frown Brow (FB). Since OpenFace~\cite{tadas2018openface,baltruvsaitis2015cross} cannot detect the AU corresponding to ``Close Eyes", we do not provide the AU accuracy for this expression.\label{tab:q1} }
\centering
\resizebox{1\linewidth}{!}{
\begin{tabular}{lcccccccccccccc}
\hline
\multirow{2}{*}{Method} & \multicolumn{7}{c}{AU Acc $\uparrow$} & \multicolumn{7}{c}{CLIP Score $\uparrow$} \\
\cmidrule(r){2-8} \cmidrule(r){9-15}
& Smile  & Angry  & Sad  & CE & RB & FB & Avg  & Smile  & Angry  & Sad   & CE  & RB  & FB & Avg  \\
\hline
DiscoFaceGAN\cite{deng2020disentangled} & 0.87 & 0.48 & 0.33 & N/A & 0.35 & 0.23 & 0.45 & 25.10 & 23.89 & 23.02 & 23.82 & 24.04 & 23.53 & 23.90 \\
AniFaceGAN\cite{wu2022anifacegan} & 0.57 & 0.43 & 0.38 & N/A & 0.60 & 0.46 & 0.49 & 24.01 & 25.94 & 24.64 & 23.98 & 24.50 & 25.06 & 24.69 \\
Morphable Diffusion\cite{chen2024morphable} & 0.46 & 0.39 & 0.52 & N/A & 0.48 & 0.48 & 0.49 & 24.28 & 25.11 & 24.29 & 25.22 & 24.75 & 25.26 & 24.82 \\
Gaussian Head Avatar\cite{xu2024gaussian} & 0.34 & 0.44 & \textbf{0.54} & N/A & 0.50 & \textbf{0.52} & 0.47 & 23.83 & 26.26 & \textbf{25.28} & 24.63 & 24.26 & 24.84 & 24.85 \\
Next3D\cite{sun2023next3d} & 0.65 & 0.41 & 0.42 & N/A & 0.49 & 0.38 & 0.47 & 24.56 & 24.61 & 23.35 & 25.93 & 24.63 & 24.61 & 24.62 \\
Diffusion-rig\cite{ding2023diffusionrig} & 0.65 & 0.38 & \textbf{0.54} & N/A & 0.66 & \textbf{0.52} & 0.55 & 24.18 & 25.76 & 24.59 & \textbf{26.30} & 24.69 & \textbf{25.40} & 25.15 \\
\hline
Ours & \textbf{0.91} & \textbf{0.49} & 0.48 & N/A & \textbf{0.68} & 0.43 & \textbf{0.60} & \textbf{25.36} & \textbf{26.52} & 24.21 & 26.13 & \textbf{24.76} & 24.68 & \textbf{25.28} \\
\hline
\end{tabular}
}
\end{table*}

\section{IMPLEMENTATION}
\subsection{Experimental Setup} 
\noindent  \textbf{Training Datasets and Details.}
During training, we utilized EMOCA~\cite{EMOCA:CVPR:2021} to extract corresponding FLAME~\cite{FLAME:SiggraphAsia2017} parameters from 2D images containing various expressions in FaceScape~\cite{yang2020facescape}, serving as the initial 3D priors. Specifically, we extract six expressions to demonstrate the editing ability of our model, i.e., ``Smile'', ``Angry'', ``Sad'', ``Close Eyes'', ``Raise Row'', and ``Frown Brow''. We train separately for each expression. Taking the expression ``Smile'' as an example, we randomly selected 5,000 latent codes from the $z$ space of the StyleGAN generator pre-trained in Next3D~\cite{sun2023next3d} to generate neural texture tri-planes. Simultaneously, previously obtained FLAME parameters for the ``Smile'' expression were randomly chosen as the driving information for generating initial images. Finally, through the optimization process guided by the text prompt ``A person who is smiling'', we obtain the final edited images. 

\noindent  \textbf{Metrics.}
We employ Fréchet Inception Distance (FID)~\cite{heusel2017gans} and Kernel Inception Distance (KID)~\cite{binkowski2018demystifying} to measure the quality of generated images. To explicitly evaluate the accuracy of fine-grained expression editing—which we define as facial modifications that capture subtle, localized muscle movements beyond coarse categorical emotion labels—we use Action Units Accuracy (AU Acc) and the CLIP Score. Unlike traditional categorical metrics that evaluate a single global emotion, AU analysis decomposes facial expressions into fundamental muscular movements based on the Facial Action Coding System (FACS). By quantitatively measuring the presence of specific Action Units (AUs) using OpenFace~\cite{tadas2018openface,baltruvsaitis2015cross},we can rigorously benchmark the structural fidelity of nuanced expressions at a micro-level. 

In practice, for each target expression, we randomly sample latent codes and reference images to generate 1,000 sample pairs. To ensure a fair evaluation across fundamentally different architectures, unconditional baselines and models trained on disparate datasets (e.g., FaceScape~\cite{yang2020facescape} and NeRSemble~\cite{kirschstein2023nersemble}) are evaluated by sampling native identities from their respective distributions. This deliberately avoids reconstruction artifacts associated with forced GAN inversion or cross-domain alignment. Crucially, to strictly rule out any selection bias caused by evaluating different visual identities, our quantitative evaluation relies entirely on identity-agnostic metrics.

Specifically, we employ the following metrics to provide a rigorous and objective benchmark:

\textbf{Action Units Accuracy (AU Acc):} We utilize OpenFace to detect localized facial muscle movements. An edit is considered successful if the generated image possesses all AUs associated with the target expression. Specifically, ``Smile'' corresponds to $AU\_06, AU\_12$; ``Angry'' to $AU\_04, AU\_05, AU\_07, AU\_23$; ``Sad'' to $AU\_01, AU\_04, AU\_15$; ``Raise Brow'' to $AU\_02$; and ``Frown Brow'' to $AU\_04$. Note that OpenFace cannot detect the corresponding AU for ``Close Eyes''.

\textbf{CLIP Score:} To evaluate semantic image-text faithfulness, we compute the cosine similarity between the embeddings of the generated image and the expression's text description in the CLIP space~\cite{taited2023CLIPScore}. Because both AU Acc and CLIP Score are fundamentally independent of the subject's underlying identity, they ensure objective measurement of fine-grained expression fidelity.

\textbf{FID and KID:} To assess overall image quality, we combine the generated samples from all expressions and compare them against a reference set of 5,000 randomly selected images from the FFHQ~\cite{karras2019style} dataset.
\begin{figure}[tb]
\centering
    \includegraphics[width=0.95\linewidth]{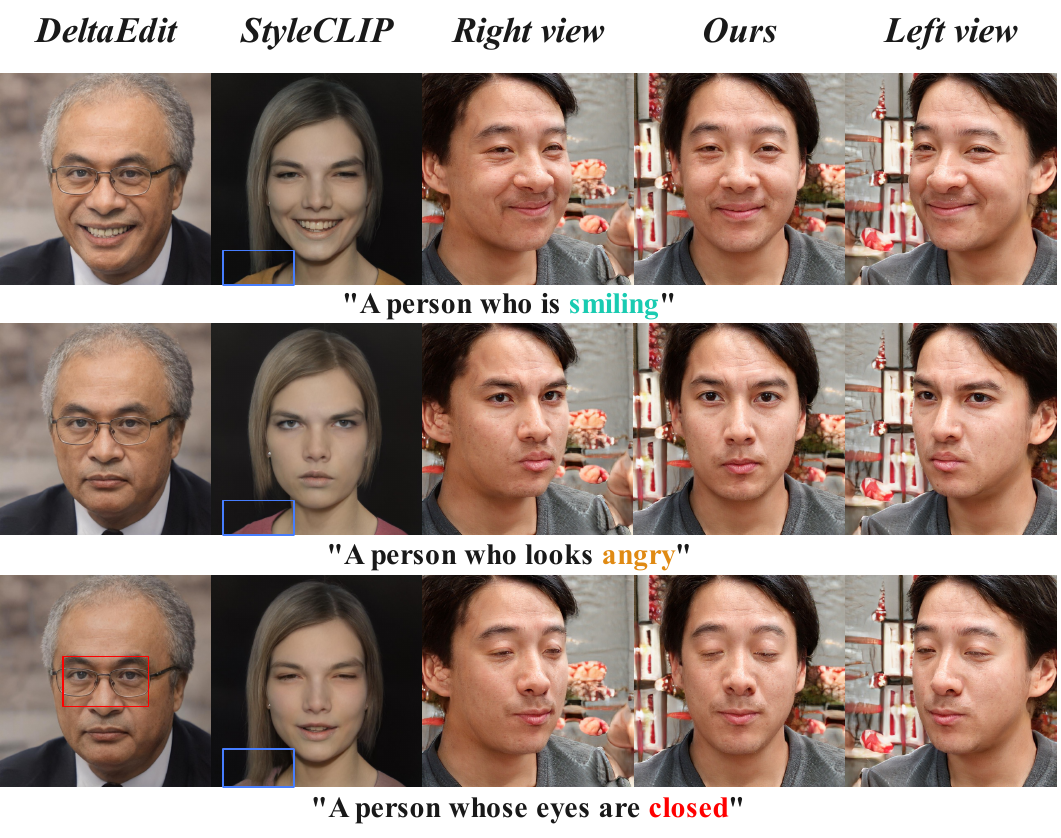}
    \caption{Comparison with the recent 2D text-guided face editing methods. We perform three expression edits on the same individual. The final three columns demonstrate view-consistent editing capabilities from our method, a feature lacking in both 2D methods, DeltaEdit and StyleCLIP, which struggle to control views.
}
    \label{figure:2D}
  \end{figure}
\noindent \textbf{Baselines.} 
We evaluate our method against several state-of-the-art baselines for 3D-aware fine-grained expression editing. Our primary comparison includes 3D-aware models compatible with single-view data: DiscoFaceGAN~\cite{deng2020disentangled}, AniFaceGAN~\cite{wu2022anifacegan}, Next3D~\cite{sun2023next3d}, and Diffusion-rig~\cite{ding2023diffusionrig}. We also compare against 2D text-guided editing methods, specifically StyleCLIP~\cite{patashnik2021styleclip} and DeltaEdit~\cite{lyu2023deltaedit}.

Additionally, we include Morphable Diffusion~\cite{chen2024morphable} and Gaussian Head Avatar~\cite{xu2024gaussian} in our comparison. Morphable Diffusion is based on a multi-view diffusion model trained on FaceScape~\cite{yang2020facescape}, while Gaussian Head Avatar utilizes 3D Gaussian Splatting (3DGS) for reconstruction on the NeRSemble~\cite{kirschstein2023nersemble} dataset. Since these methods rely on extensive multi-view data and are difficult to adapt to the single-view FFHQ~\cite{karras2019style} dataset, we focus solely on comparing their expression editing performance using their officially released code and pretrained models.

For the remaining baselines, all results are attained using their released code and pretrained models on the FFHQ dataset, except for StyleCLIP, which required retraining using default hyperparameters as no pretrained model was available. For Diffusion-rig, we further fine-tuned the model for each identity using Next3D-generated images with diverse expressions to support the required image generation.
\vspace{-12pt}
\subsection{Qualitative Evaluation}
\subsubsection{3D Methods}
Fig.~\ref{figure:3D} illustrates the comparative results between our approach and the 3D-aware animatable baselines. Overall, our method more accurately captures the fine-grained facial expression features in the reference images and generates high-quality images that are consistent with the target facial expression text descriptions. Specifically, Next3D can tolerate the changes in topology in the expression control by adapting surface deformation into a continuous volumetric representation. Diffusion-rig can extract the expression coefficients of the reference image using a DECA~\cite{DECA:Siggraph2021}, then generate the corresponding mesh, which is subsequently rendered into an albedo map used as a condition for the diffusion model.
\begin{table*}[t]
\centering
\caption{Quantitative results of image quality using FID and KID.}
\label{tab:q2}
	\resizebox{0.85\linewidth}{!}{
\begin{tabular}{ccccccccc}
\toprule
& StyleCLIP~\cite{patashnik2021styleclip} & DeltaEdit~\cite{lyu2023deltaedit} & DiscoFaceGAN~\cite{deng2020disentangled} & AniFaceGAN~\cite{wu2022anifacegan} & Next3D~\cite{sun2023next3d}&Diffusion-rig~\cite{ding2023diffusionrig}& Ours  \\  \midrule                        
FID $\downarrow$ & 35.53  & 45.78    & 58.95     & 51.00   & 32.95 & 31.26 &  \textbf{30.71 }\\
KID $\downarrow$ & 0.021  & 0.024 & 0.040 & 0.035 & 0.019 & 0.024 &  \textbf{0.018 } \\
\bottomrule
\end{tabular}}
\vspace{-10pt}
\end{table*}
Consequently, ours, Diffusion-rig and Next3D can successfully generate the ``Close Eyes'' expression. In contrast, DiscoFaceGAN and AniFaceGAN struggle to match this challenging expression, as depicted in Fig.~\ref{figure:3D} (Blue Box). Similarly, although Morphable Diffusion and Gaussian Head Avatar align well with the majority of target expressions, they perform poorly on challenging expressions such as ``Close Eyes''. Specifically, generating challenging expressions like ``Close Eyes'' exposes two primary bottlenecks in existing methods: data bias in standard 3D priors (e.g., DECA), which are dominated by open-eyed subjects, and uncoordinated updates between eyelid geometry and eye textures, leading to `ghosting' artifacts. Our approach overcomes these issues by leveraging EMOCA for a more accurate geometric initialization. Crucially, our Dual Mappers synergistically coordinate topological mesh deformations with UV texture updates, while the CLIP loss provides semantic compensation for the scarcity of closed-eye 3D data. This ensures robust, natural closed-eye generation without texture-mesh mismatches. 

Furthermore, since DiscoFaceGAN and Diffusion-rig do not explicitly utilize 3DMM information to model the entire 3D representation corresponding to the generated images, they encounter view-inconsistency issues. Although Next3D and Diffusion-rig can match most target expressions, their ability to capture fine-grained expressions is limited. Specifically, as shown in Fig.~\ref{figure:3D} (Red box) and zoomed results in Fig.~\ref{figure:zoom-in}, Next3D and Diffusion-rig do not align well with both the reference image and the given expression category simultaneously. For example, they may exhibit ``Smile'' characteristics in ``Sad'' expressions, shown in Fig.~\ref{figure:3D} ``Sad'', or they may fail to represent some expressions like ``Raise brow'', shown in Fig.~\ref{figure:zoom-in}.

Overall, our method not only achieves ``challenging expression'' editing by successfully modeling topology changes (e.g., Close eyes) but also accurately generates the fine-grained expression features in the reference images and aligns more closely with the textual description of the target expression.

\subsubsection{2D Methods}
To assess the alignment between the edited expressions and the given text prompts achieved by our method, we compare the editing quality with StyleCLIP~\cite{patashnik2021styleclip} and DeltaEdit~\cite{lyu2023deltaedit} under the given text descriptions, such as ``A person who is smiling''. As shown in Fig.~\ref{figure:2D}, generated images from StyleCLIP can match the corresponding text descriptions well, but they have poor background preservation (the clothes color has been changed in the blue box). On the other hand, although DeltaEdit can maintain good background preservation, it does not match well with some expressions such as ``eye closed'' (Red box). In contrast, our method not only achieves better preservation of background but also produces edited expressions closer to the corresponding text descriptions. Moreover, our method can generate high-quality images from novel views and maintain good 3D consistency. More results can be found in the Demo Video.

\begin{table}[t]
\caption{Quantitative results on the fine-grained expression editing results using CLIP Score. The used expressions are Smile, Angry, Sad, Close Eyes (CE), Raise Brow (RB), and Frown Brow (FB). }
\label{tab:q4}
\centering
\setlength\tabcolsep{2pt}
\resizebox{0.95\linewidth}{!}{
\begin{tabular}{lccccccc}
\toprule
\multirow{2}{*}{Method} & \multicolumn{7}{c}{CLIP Score $\uparrow$} \\
\cmidrule(r){2-8}
 & Smile  & Angry  & Sad & CE & RB  & FB  & Avg   \\
\midrule
DeltaEdit~\cite{lyu2023deltaedit} & \textbf{25.81} & 23.69 & 23.26 & 23.41  & 23.51 & 22.68 & 23.73 \\
StyleCLIP~\cite{patashnik2021styleclip} & 25.15  & \textbf{26.48}  & \textbf{25.69}  & 23.59 & 23.67 & 23.27  & 24.64  \\
\hline
Ours & 24.70 & 26.27 & 23.96& \textbf{26.16} & \textbf{24.66} & \textbf{25.06}  & \textbf{25.14} \\
\bottomrule
\end{tabular}}
\end{table}
\vspace{-12pt}
\subsection{Quantitative Evaluation}
Table~\ref{tab:q1} presents the comparison results of our method with 3D-aware animatable baselines in terms of AU Acc and CLIP Score. Our method achieves the best scores in most expressions. It is observed that DiscoFaceGAN performs well on the "Smile" expression due to its training set containing many "Smile"-related images, but struggles with other expressions. Its limited ability to extract and generate fine-grained expression features leads to generating mostly "Smile"-related images. For other expressions, our method achieves higher average AU Acc and CLIP Score, indicating a better ability to extract fine-grained expression features from the reference image and maintain consistency with the expression text description. Additionally, for the "Close Eyes" expression, OpenFace~\cite{tadas2018openface,baltruvsaitis2015cross} fails to detect the corresponding AU, but the CLIP Score shows that our method can effectively handle this more challenging expression editing task.

Table~\ref{tab:q2} presents the visual quality evaluation results of our method compared to all baselines. Our method demonstrates significant improvements in FID and KID compared to 3D methods AniFaceGAN and DiscoFaceGAN, as well as 2D methods StyleCLIP and DeltaEdit. It is noted that, except for Next3D, all 3D methods generate results at $256^2$ resolution, while the 2D methods generate results at $1024^2$ resolution. Therefore, we first select images at $1024^2$ resolution from the FFHQ dataset and then resize them to $256^2$ and $512^2$ resolutions for comparison. Our method also exhibits certain improvements compared to Next3D and Diffusion-rig, indicating that our proposed modules and optimization methods do not compromise the generation quality. We note that Morphable Diffusion~\cite{chen2024morphable} and Gaussian Head Avatar~\cite{xu2024gaussian} are excluded from Table~\ref{tab:q2}. Since they utilize FaceScape and NeRSemble, respectively, the inherent dataset discrepancy heavily impacts FID and KID scores. Consequently, we limit the comparison with these methods to AU Accuracy and CLIP Score to ensure a fair evaluation.

In Table~\ref{tab:q4}, we compare the text-guided 2D face editing methods DeltaEdit~\cite{lyu2023deltaedit} and StyleCLIP~\cite{patashnik2021styleclip} in terms of CLIP Score under given textual descriptions. Our method performs the best on most expressions, especially on challenging ones like ``Close eyes''. This indicates that our method outperforms 2D approaches by better aligning with corresponding textual descriptions.

\begin{table*}[t]
\setlength\tabcolsep{3pt}
\caption{Ablation study on fine-grained expression editing quality and image quality. \label{tab:abation1} }

\centering
\resizebox{0.95\linewidth}{!}{
\begin{tabular}{lllllllllllllll|cc}
\toprule
\multirow{2}{*}{Method} & \multicolumn{7}{c}{AU Acc $\uparrow$} & \multicolumn{7}{c}{CLIP Score $\uparrow$} & \multicolumn{2}{c}{FID\&KID $\downarrow$} \\
\cmidrule(r){2-8} \cmidrule(r){9-15} \cmidrule(r){16-17}
 & Smile  & Angry  & Sad & CE & RB  & FB  & Avg & Smile & Angry & Sad & CE & RB & FB & Avg & FID & KID \\
\midrule
\textit{w/o} $\mathcal{L}_\mathrm{ID}$ & 0.99 & 0.48 & 0.81 & N/A & 0.92 & 0.21 & 0.68 & 27.30 & 28.80 & 26.07 & 27.94 & 26.96 & 29.29 & 27.73 & 43.42 & 0.027 \\
\textit{w/o Projection} & 0.75 & 0.46 & 0.49 & N/A & 0.68 & 0.67 & 0.61 & 24.89 & 25.77 & 23.98 & 26.18 & 24.85 & 25.62 & 25.22 & 30.38 & 0.018 \\
\textit{w/o Cross-Attention} & 0.83 & 0.56 & 0.48 & N/A & 0.68 & 0.62 & 0.63 & 25.34 & 26.43 & 23.94 & 26.28 & 24.76 & 25.56 & 25.38 & 30.93 & 0.018 \\
\textit{w/o EMOCA} & 0.70 & 0.48 & 0.52 & N/A & 0.64 & 0.47 & 0.56 & 24.79 & 26.44 & 24.44 & 26.49 & 24.80 & 25.52 & 25.41 & 33.47 & 0.019 \\
\hline
Full Model & 0.90 & 0.49 & 0.56 & N/A & 0.73 & 0.56 & 0.65 & 25.12 & 26.38 & 24.67 & 26.10 & 24.72 & 24.96 & 25.33 & 30.13 & 0.017 \\
\bottomrule
\end{tabular}}
\vspace{-5pt}
\end{table*}

\begin{figure}[tb]
\centering
    \includegraphics[width=1\linewidth]{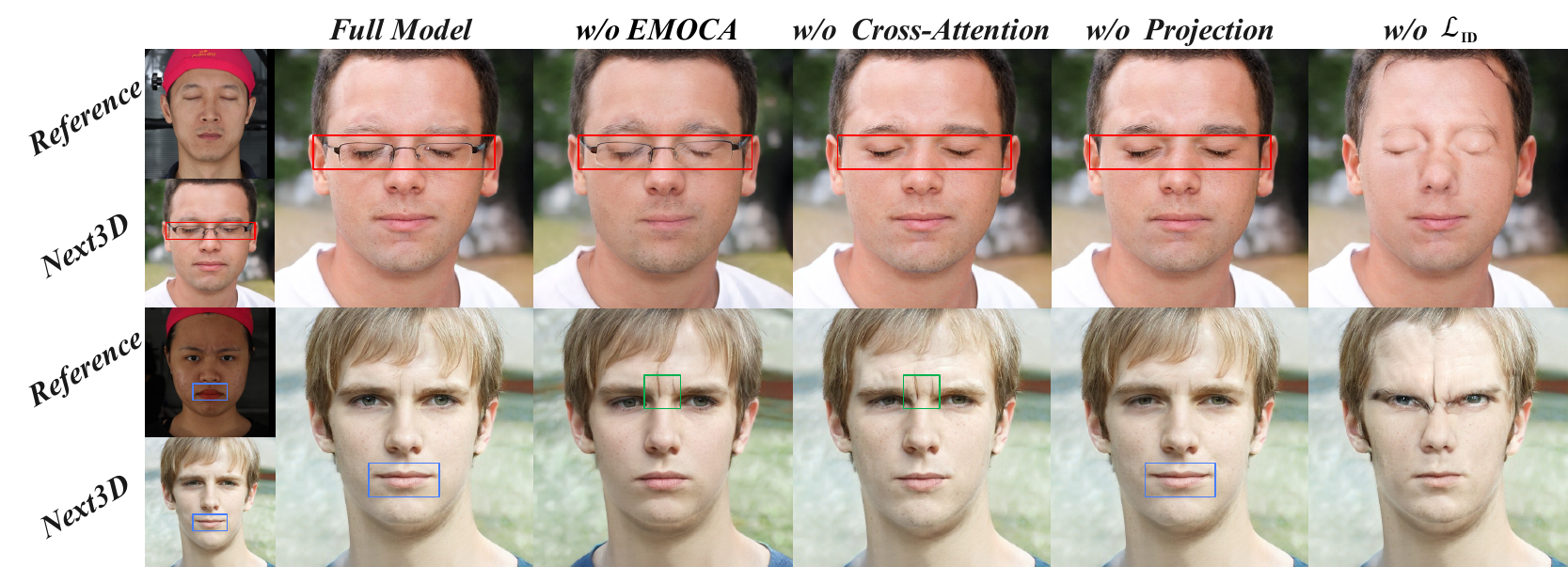}
    \caption{The qualitative ablation experiments examine the main components of our method, including the removal of identity loss (\textit{w/o} $\mathcal{L}_{\mathrm{ID}}$), Subspace Projection (\textit{w/o Projection}), Cross-Attention (\textit{w/o Cross-Attention}) and EMOCA (\textit{w/o EMOCA}). We present the results of expression editing on the ``Close Eyes'' and ``Angry'' expressions, for each ablation.
}
    \label{figure:Ablation}
  \end{figure}
\begin{table*}[ht]
    \centering
    \setlength\tabcolsep{2pt}
            \caption{Quantitative results for background preservation.
}
    \resizebox{\linewidth}{!}{
    \begin{tabular}{c|c|c|c|c|c|c|c}
    \hline
    Metric & DiscoFaceGAN~\cite{deng2020disentangled}& AniFaceGAN~\cite{wu2022anifacegan}  & Morphable Diffusion~\cite{chen2024morphable} & Gaussian Head Avatar~\cite{xu2024gaussian} & Next3D~\cite{sun2023next3d} & Diffusion-rig~\cite{ding2023diffusionrig}& Ours 
    \\ 
    \hline
         SSIM $\uparrow$ & 0.8457 & 0.9996 & 0.9911 & 0.9998 & 0.9998 & 0.9998  &  0.9996 \\
    \hline
    \end{tabular}
    }
    \label{tab:quanv2}
    \vspace{-10pt}
\end{table*}
\vspace{-12pt}
\subsection{Ablation Study}

We also conducted an evaluation to assess the impact of various design choices within our method, including the identity loss \(\mathcal{L}_{\mathrm{ID}}\), Subspace Projection, Cross-Attention Module, and EMOCA. The qualitative results depicted in Fig.~\ref{figure:Ablation} illustrate that our full model achieves greater precision in fine-grained expression editing compared to all ablations.
Notably, the altered face identity in the absence of \(\mathcal{L}_{\mathrm{ID}}\) underscores the effectiveness of \(\mathcal{L}_{\mathrm{ID}}\) in preserving identity. Similarly, the absence of EMOCA leads to a noticeable reduction in the accuracy of expression editing, as highlighted in Fig.~\ref{figure:Ablation} (Red Box). Without EMOCA, the fine-grained facial movements are not accurately captured, resulting in less expressive and misaligned outputs. Moreover, both the absence of Subspace Projection and the Cross-Attention Module result in excessive editing problems. Particularly, the absence of Subspace Projection causes unintended attribute alterations, as seen in Fig.~\ref{figure:Ablation} (Red Box). The blue box highlights the improved alignment of subtle facial features, such as the curvature of the mouth corners, achieved by incorporating Subspace Projection. Additionally, the green box demonstrates that the inclusion of the Cross-Attention Module reduces the generation of artifacts, ensuring smoother and more coherent editing results.

Table~\ref{tab:abation1} demonstrates that our method outperforms all other ablations in terms of AU Acc, except for the \textit{w/o} $\mathcal{L}_\mathrm{ID}$ variant. In the absence of $\mathcal{L}_\mathrm{ID}$, the model excessively edits the expression under CLIP guidance, resulting in generated images that exhibit stronger features of the target expression, but struggle to maintain identity consistency and image quality. This is reflected in both the FID and KID scores. In terms of CLIP Score, the full model performs similarly to the other ablations, with \textit{w/o} $\mathcal{L}_\mathrm{ID}$ still achieving the best results. For FID and KID, the Full model achieves the best performance. In conclusion, when considering both the accuracy of expression editing and the quality of generated images, the Full Model exhibits the best performance.
\begin{figure}[t]
\centering
\includegraphics[width=1.0\linewidth]{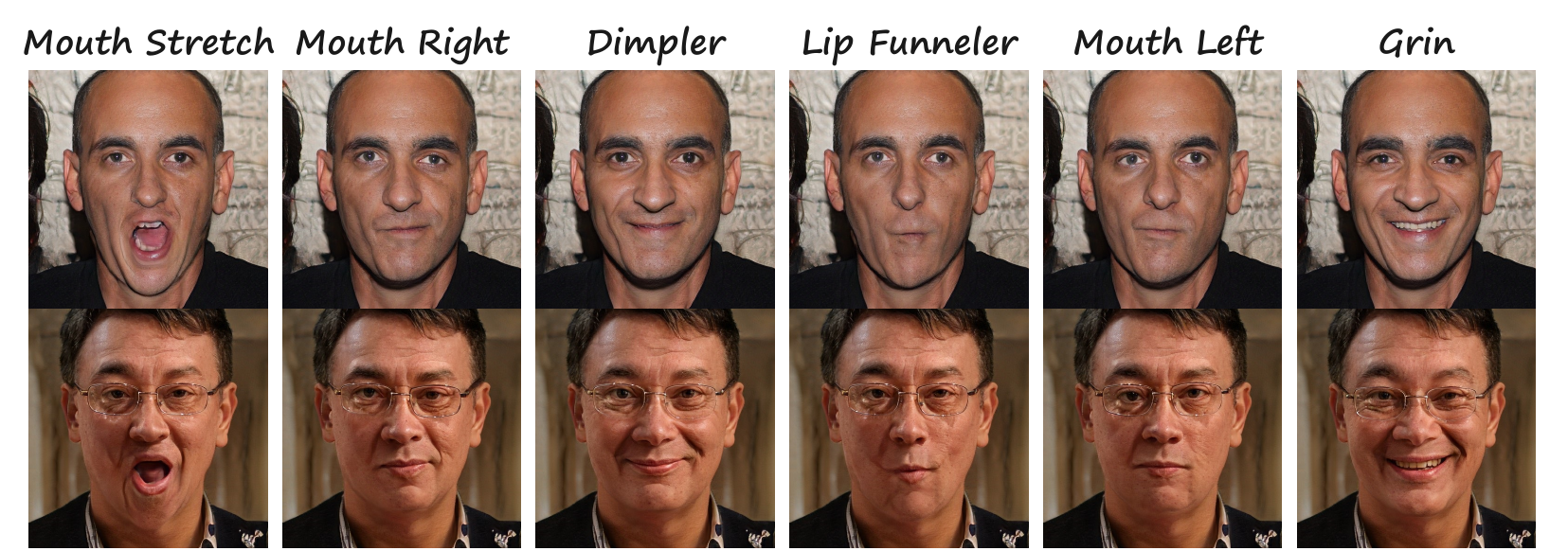}
\caption{Illustration of new expression editing results using the proposed method, including ``Mouth Stretch", ``Mouth Right" ``Dimpler", ``Lip Funneler", ``Mouth Left" and ``Grin". The edited images demonstrate the model's ability to generate diverse fine-grained expressions while maintaining high visual quality.}
\label{fig:memo}
\end{figure}
\begin{figure}[t]
\centering
\includegraphics[width=1.0\linewidth]{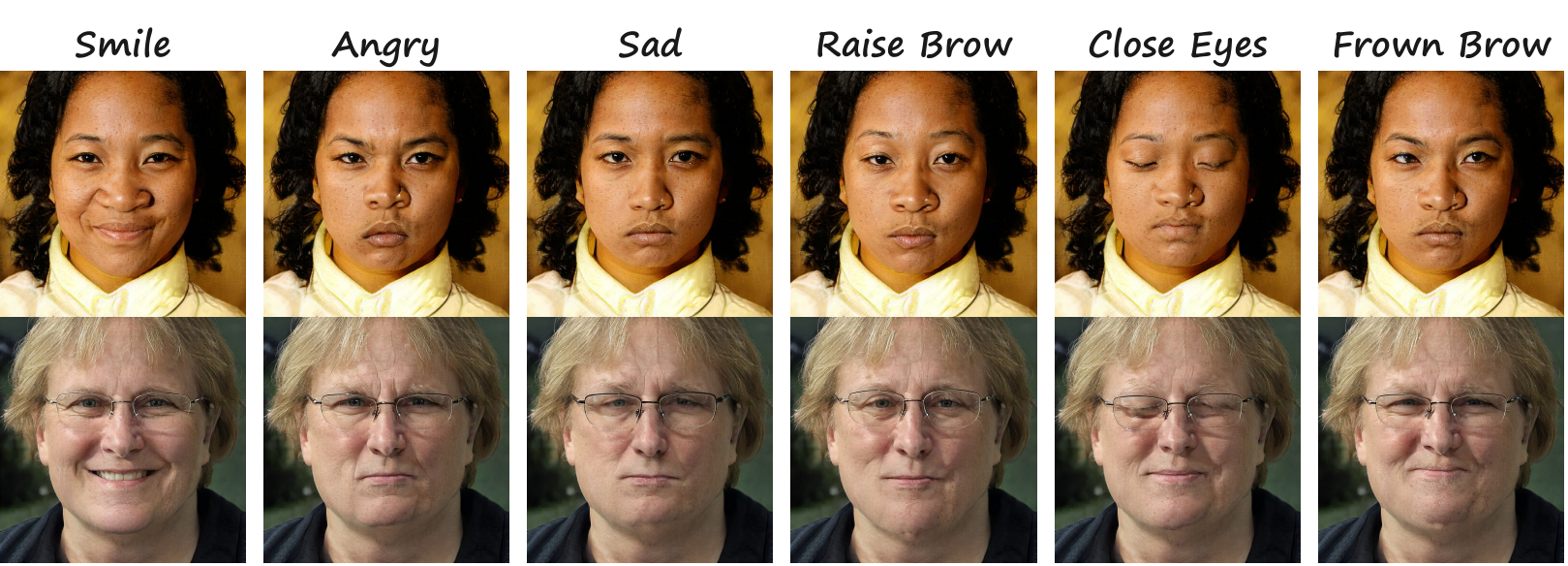}
\caption{Expression editing results for the same individual. Each row corresponds to a different subject, and each column represents a target expression, including ``Smile", ``Angry", ``Sad", ``Raise Brow", ``Close Eyes", and ``Frown Brow". The results demonstrate the ability of the proposed method to modify expressions while preserving the identity and overall image quality.}
\label{fig:id2}
\end{figure}
\section{Identity consistency maintenance and results for additional expressions.}
We provide more expression editing results in Fig.~\ref{fig:memo} and Fig.~\ref{fig:id2}. Fig.~\ref{fig:memo} shows new expression editing which demonstrates that our method improves the editing precision of Next3D for subtle expressions while supporting common expressions. Both figures illustrate multiple expression editing for the same identity. Notably, the identity remains well-preserved. Moreover, Table~\ref{tab:quanv2} provides a quantitative evaluation of the preservation of non-target regions (backgrounds), with the high scores highlighting the effectiveness of our method in maintaining identity.
\begin{figure}[t]
\centering
\includegraphics[width=0.9\linewidth]{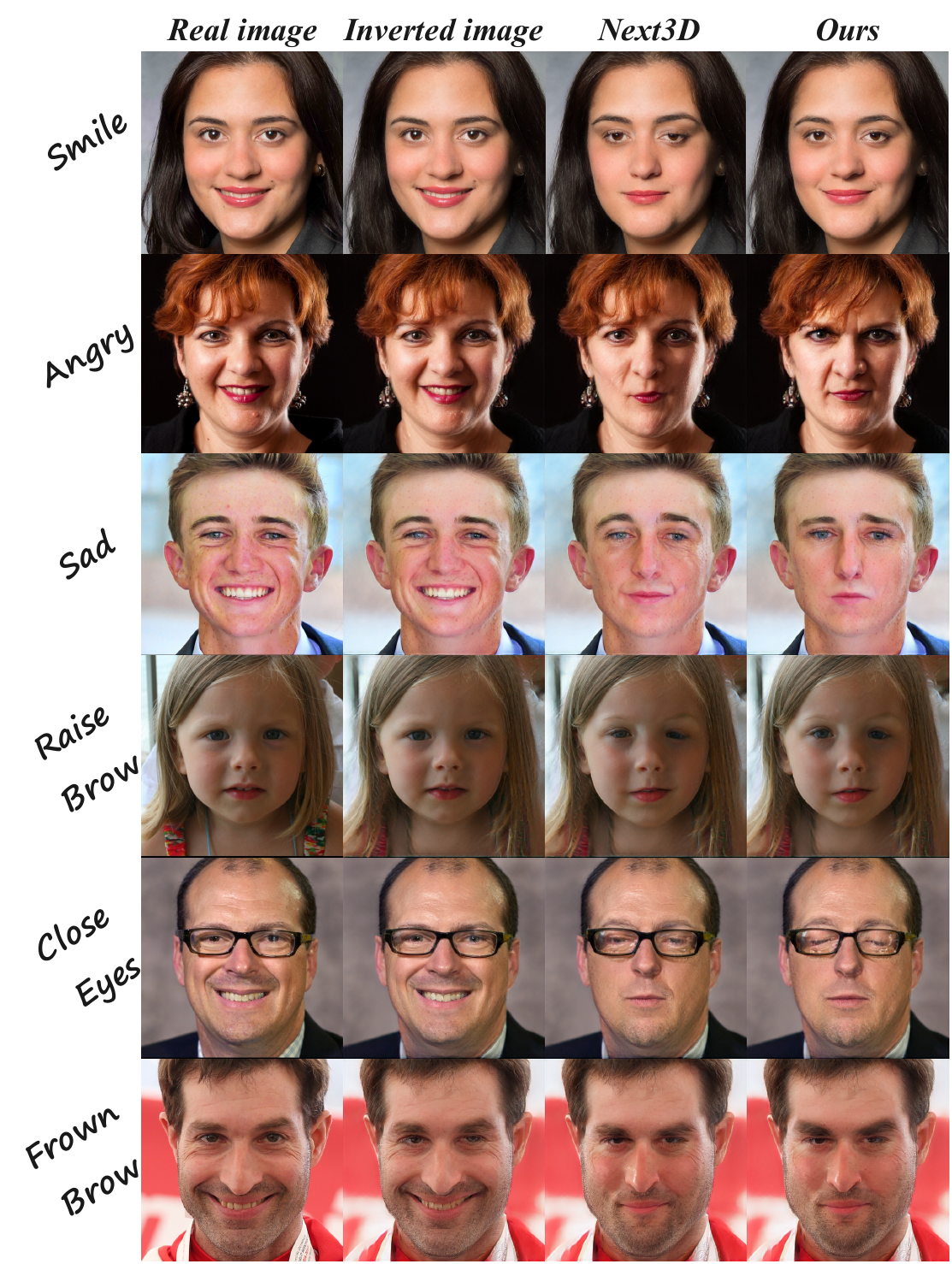}
\caption{Comparison results of fine-grained expression editing on real images: real images, inverted images, Next3D outputs, and results from the proposed method across six expressions (``Smile'', ``Angry'', ``Sad'', ``Raise Brow'', ``Close Eyes'', and ``Frown Brow'').}
\label{fig:inversion}
\end{figure}
\section{Inversion and Editing Results}
To demonstrate the effectiveness of our method on real-world images, we randomly selected several images from the FFHQ dataset~\cite{karras2019style} and utilized the 3D GAN inversion technique described in~\cite{ko20233d}. The inverted images were then edited to modify their facial expressions, with the results presented in Fig.~\ref{fig:inversion}. As shown, the inverted images maintain a high level of fidelity, closely resembling the original real images. By applying our expression editing approach, we successfully generated images that not only retain high visual quality but also depict facial expressions that align more accurately with the corresponding text descriptions. This highlights the robustness of our method in handling real-image scenarios and demonstrates its ability to preserve essential image details while achieving expressive edits.

\begin{figure}[t]
\centering
\includegraphics[width=0.9\linewidth]{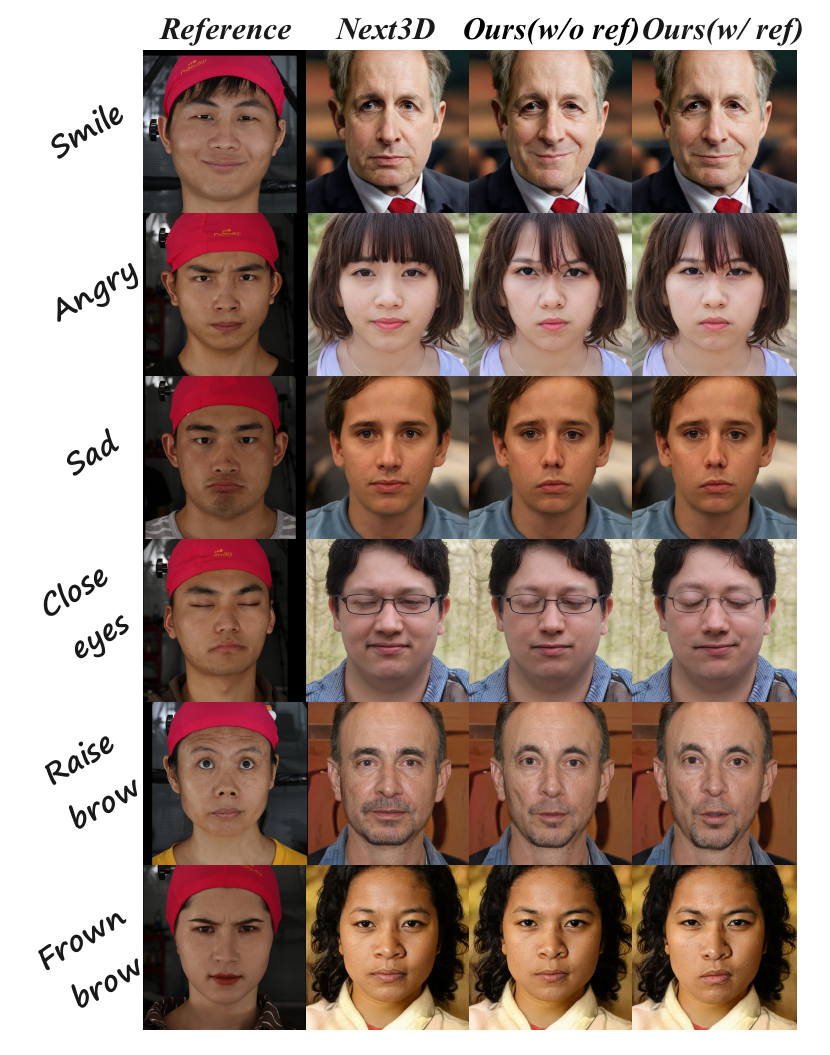}
\caption{Comparison results under different training setups: training without target expression data versus training initialized with target expression codes, evaluated across six expressions (``Smile'', ``Angry'', ``Sad'', ``Raise Brow'', ``Close Eyes'', and ``Frown Brow'').}
\label{fig:without_ref}
\end{figure}

\section{Influence of Expression Data}
In practical applications, obtaining large-scale, high-quality facial image datasets with diverse fine-grained expressions poses significant challenges. This makes it impractical to adopt methods that require training with a large number of images matching the target expressions for initializing expression codes. In our method, we use target expression images to initialize expression parameters and CLIP loss for optimization, leveraging textual descriptions to guide the generated expressions and align them with the intended semantics. Specifically, we present the results in Fig.~\ref{fig:without_ref}, comparing the training initialized with neutral expressions (\textit{w/o ref}) against those using reference images of the target expressions (\textit{w/ ref}). While the \textit{w/o ref} method is not as effective as the \textit{w/ ref} approach in terms of editing performance, it still achieves noticeable editing results, particularly for expressions like Smile, Angle, and Sad. This improvement can be attributed to the CLIP loss, which plays a pivotal role in aligning the generated expressions with the intended textual descriptions. 

\section{View consistency analysis of Diffusion based and 3D-aware GAN based methods}
\begin{figure}[ht]
\centering
\includegraphics[width=0.90\linewidth]{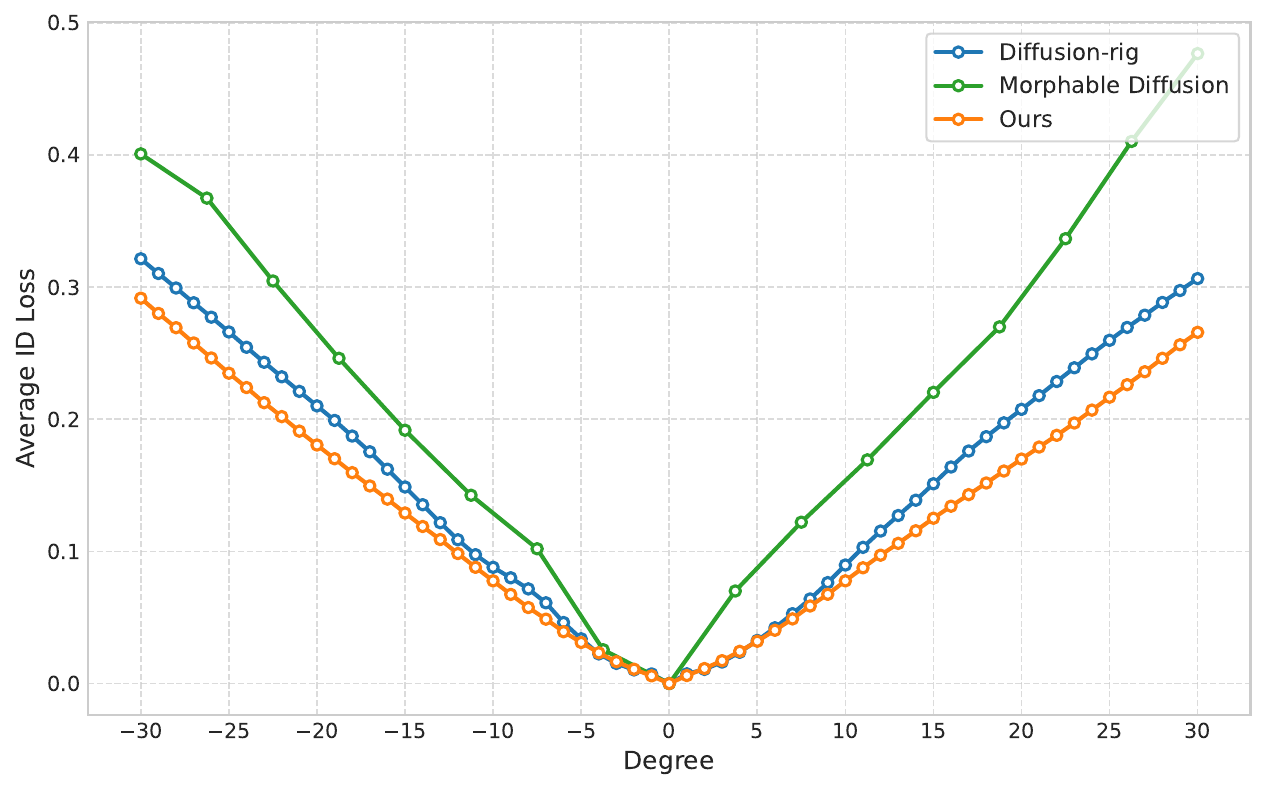}
\caption{The view consistency across different viewpoints is evaluated by calculating \(\mathcal{L}_{\mathrm{ID}}\) between images generated by rotating the head model by 1-30 degrees from the frontal view. Our method demonstrates a smoother trend compared to Diffusion-rig and Morphable Diffusion (evaluated via uniform sampling from $-30^\circ$ to $30^\circ$ due to its 16-image generation limit).}
\label{fig:3D-consis}
\end{figure}
\begin{figure}[ht]
\centering
\includegraphics[width=1.0\linewidth]{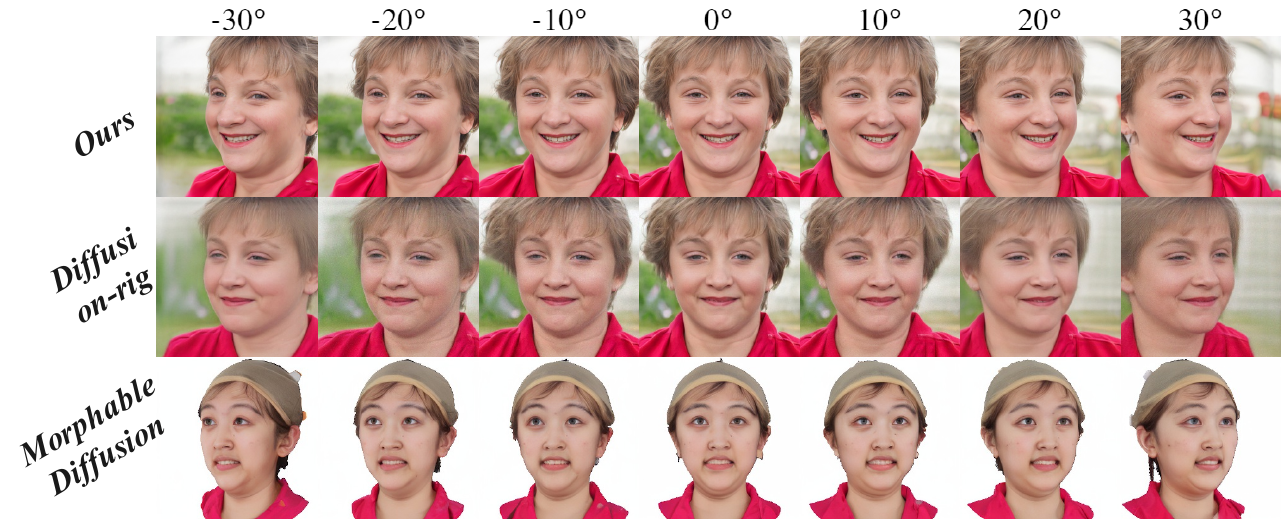}
\caption{The images generated from different viewpoints demonstrate that our method maintains superior view consistency, particularly in the generation of hair, compared to Diffusion-rig.
}
\label{fig:3D_consis}
\end{figure}

As shown in Fig.~\ref{fig:3D-consis} and Fig.~\ref{fig:3D_consis}, diffusion-based methods lack explicit 3D head modeling, leading to artifacts in view consistency. For instance, Diffusion-rig relies on conditional image generation, and Morphable Diffusion, despite employing a multi-view diffusion architecture, still generates novel views without a consistent 3D geometry. Consequently, they exhibit inconsistencies when the viewing angle changes. In contrast, 3D-aware GAN-based methods like ours model the head utilizing neural radiance fields, ensuring strict geometric consistency across different rendered viewpoints. As a result, our method maintains better view consistency. 
\begin{table}[tp]
    \centering
    \caption{Comparison of inference time (in seconds) per run across different methods. Our method achieves competitive efficiency compared to GAN-based methods while significantly outperforming diffusion-based baselines.}
    \label{tab:inference_time}
    \resizebox{0.75\linewidth}{!}{%
        \begin{tabular}{l|c}
            \toprule
            \textbf{Methods} & \textbf{Inference Time (s)} \\
            \midrule
            DiscoFaceGAN~\cite{deng2020disentangled} & 0.62 \\
            AnifaceGAN~\cite{wu2022anifacegan} & 0.95 \\
            Next3D~\cite{sun2023next3d} & 0.52 \\
            Gaussian Head Avatar~\cite{xu2024gaussian} & 0.42 \\
            \midrule
            Diffusion-rig~\cite{ding2023diffusionrig}& 1.58 \\
            Morphable Diffusion~\cite{chen2024morphable}& 1.26 \\
            \textbf{Ours}         & \textbf{0.71} \\
            \bottomrule
        \end{tabular}%
    }
\end{table}

\section{Computational Complexity Analysis}
We mainly evaluate the computational efficiency of our proposed method in terms inference time. All experiments were conducted on a workstation equipped with a single NVIDIA RTX 4090 GPU (24GB VRAM).

Our approach trains a dedicated \textit{Dual Mapper} for each specific target expression. The training process for refining a Dual Mapper typically requires approximately 14 hours. While our method involves an optimization-based training strategy, which is computationally intensive, this is a one-time cost. Once the Dual Mapper is trained, no further optimization is required during the inference stage. 

Unlike optimization-based editing methods that require time-consuming iterations during generation, our method achieves optimization-free inference. As shown in Table~\ref{tab:inference_time}, we compare the inference time of our method against several approaches. While the 3DGS-based reconstruction method, Gaussian Head Avatar, achieves the fastest inference speed, it requires two days of training for each individual subject. Our method requires only 0.71 seconds per run. This is significantly faster than diffusion-based editing methods such as Diffusion-rig (1.58s) and Morphable Diffusion (1.26s), and is comparable to GAN-based approaches like DiscoFaceGAN (0.62s) and Next3D (0.52s), striking a favorable balance between generation quality and computational efficiency.

\section{Sensitivity Analysis of $\lambda_{\mathrm{ID}}$:}
\begin{figure}
    \centering
  \includegraphics[width=0.9\linewidth]{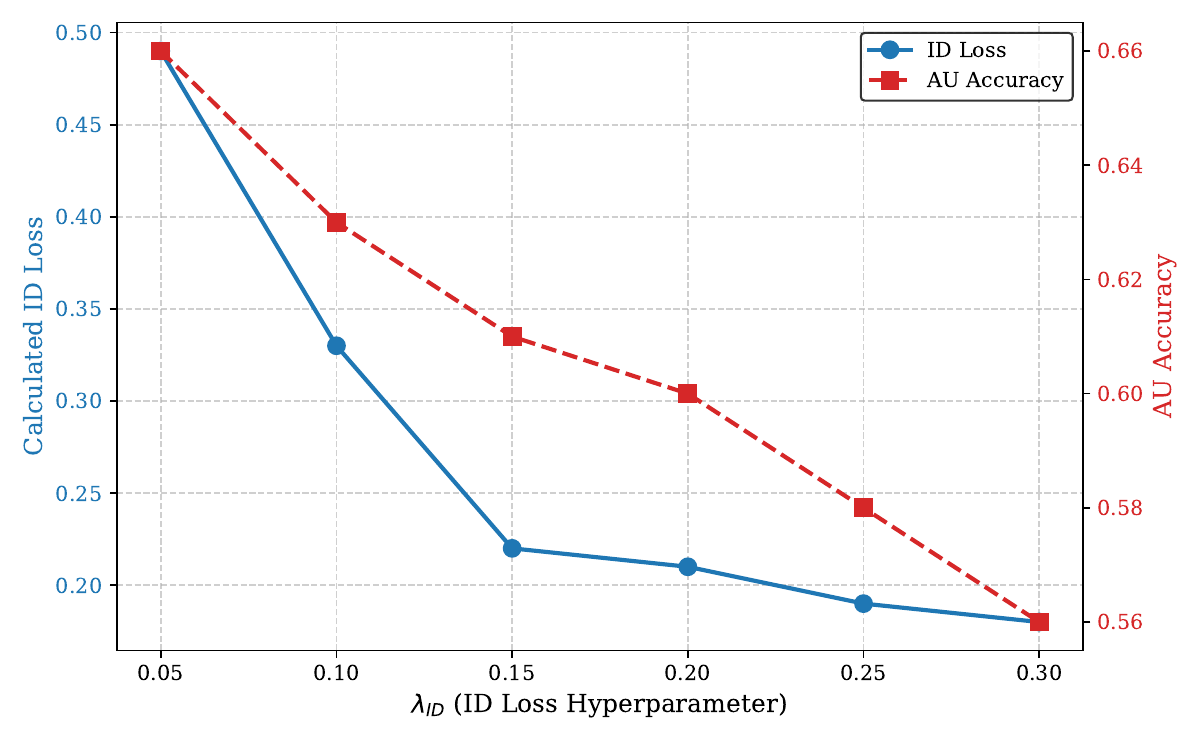}
  \caption{Sensitivity Analysis of the Identity Regularization Weight ($\lambda_{\mathrm{ID}}$).}
  \label{fig:id_loss_vs_au_acc}
\end{figure}
We evaluated the network's performance while varying $\lambda_{\mathrm{ID}}$ from 0.05 to 0.30 in intervals of 0.05. The analysis tracks two key metrics: Calculated ID Loss (measuring identity preservation, where lower is better) and AU Accuracy (measuring expression editing precision). As illustrated in Fig.~\ref{fig:id_loss_vs_au_acc}, increasing $\lambda_{\mathrm{ID}}$ effectively reduces the Calculated ID Loss, indicating stronger identity preservation. However, this comes at the cost of expression expressiveness, evidenced by the steady decline in AU Accuracy as the parameter increases. We observe a sharp improvement in identity preservation (a steep drop in ID loss) as $\lambda_{\mathrm{ID}}$ increases from 0.05 to 0.15. Beyond this point, the identity loss begins to stabilize. The range between 0.15 and 0.20 represents a ``sweet spot'' where identity features are robustly maintained without overly sacrificing the target expression's accuracy. To ensure the high-fidelity retention of the source identity while maintaining competitive expression generation capabilities, we empirically selected $\lambda_{\mathrm{ID}} = 0.20$ as our final hyperparameter setting for all primary experiments.

\section{Compositional Expression Editing}

\begin{figure}[t]
  \includegraphics[width=0.95\linewidth]{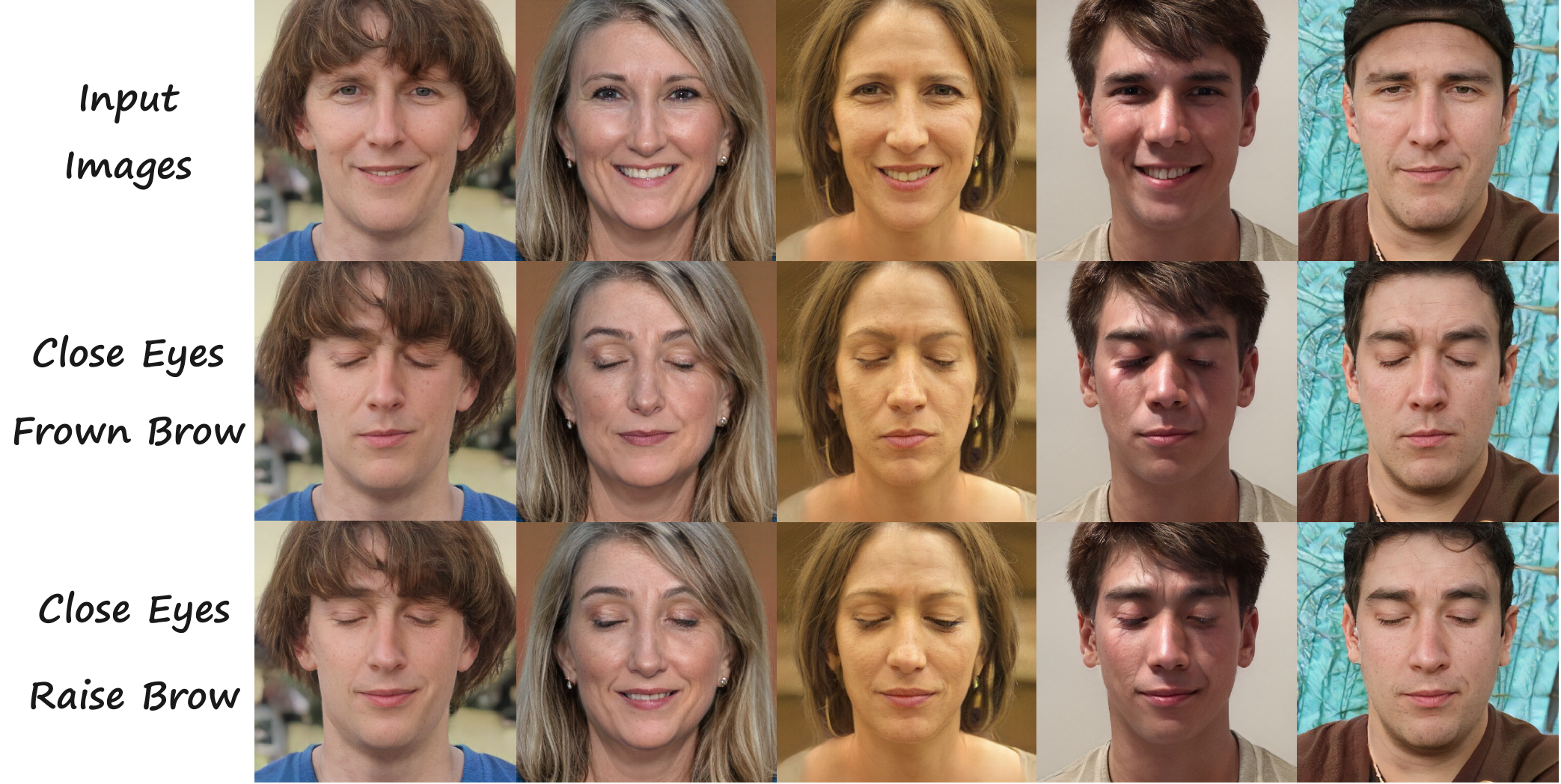}
  \caption{Qualitative results of compositional expression editing demonstrate our model's ability to seamlessly combine multiple distinct facial actions, such as ``Close Eyes'' alongside ``Frown Brow'' or ``Raise Brow'', applied to various input images.}
  \label{fig:two_emo}
\end{figure}

To further evaluate our model's capabilities, we investigate its performance in compositional expression generation by combining multiple distinct facial actions. As illustrated in Fig.~\ref{fig:two_emo}, we successfully combined the action ``Close Eyes'' with either ``Frown Brow'' or ``Raise Brow'', applying them simultaneously to various input images. The visual results clearly demonstrate that our model handles these composite edits seamlessly. The distinct facial actions—such as the closing of the eyelids and the vertical or inward movement of the eyebrows—are synthesized together with high fidelity. The geometric deformations in these adjacent regions do not conflict or cause unnatural artifacts, confirming that our framework can effectively merge multiple, distinct expression instructions into a single, cohesive output.
\section{Limitations and Discussion}
Our proposed editing method may not be effective for all generated images. For instance, when the initial expression of some images differs significantly from the reference image, or when the reference image does not clearly exhibit the expression corresponding to the text description, artifacts may appear in the generated images, or they may fail to reflect the intended expression. Additionally, challenging expressions such as ``Close eyes'' may introduce more artifacts. Fine-tuning the pretrained Next3D model with a more diverse dataset of expressions, and using more accurate reference images for driving, may reduce the occurrence of these situations.
Large Generative Models are one type of method requiring intensive computational resources in general. Compared to the methods training from scratch, our method takes a pretrained 3DGAN as the backbone, thus saving largely computational resources in the training stage. What we want to emphasize is that our method focuses on the expression editing problem. The speed of image editing mainly depends on the rendering progress of the NeRF module used, as our dual mappers are small networks. In addition, using better 3D representations, such as 3D Gaussian instead of NeRF, may improve rendering speed in the future. 

\section{Conclusion}
We have introduced a method that directly utilizes the pretrained 3D-Aware Animitable GAN model for fine-grained expression editing tasks. Our approach involves the incorporation of the Dual Mappers module into the pretrained model, allowing for the simultaneous refinement of the latent space of UV texture and the expression space of the driven mesh. Additionally, we have proposed a Text-Guided Optimization method, which employs a CLIP-based objective function with expression text prompts as targets. Furthermore, we have integrated a SubSpace Projection mechanism to provide more precise control over fine-grained expressions. These advancements pave the way for more effective and nuanced expression editing capabilities within the realm of computer graphics and animation. In future work, we aim to explore additional techniques for further enhancing the expressiveness and realism of expression editing. This may involve investigating novel approaches for refining the alignment between text prompts and edited expressions, as well as exploring the integration of multimodal inputs, such as audio cues or physiological signals, to enrich the editing process.
\let\oldbibliography\bibliography
\renewcommand{\bibliography}[1]{
  \setlength{\itemsep}{0pt} 
  \oldbibliography{#1}
}
\bibliographystyle{IEEEtran}
\bibliography{main}

\vspace{-15pt}
\begin{IEEEbiography}[{\raisebox{0pt}{\includegraphics[width=1in,height=1.25in,clip]{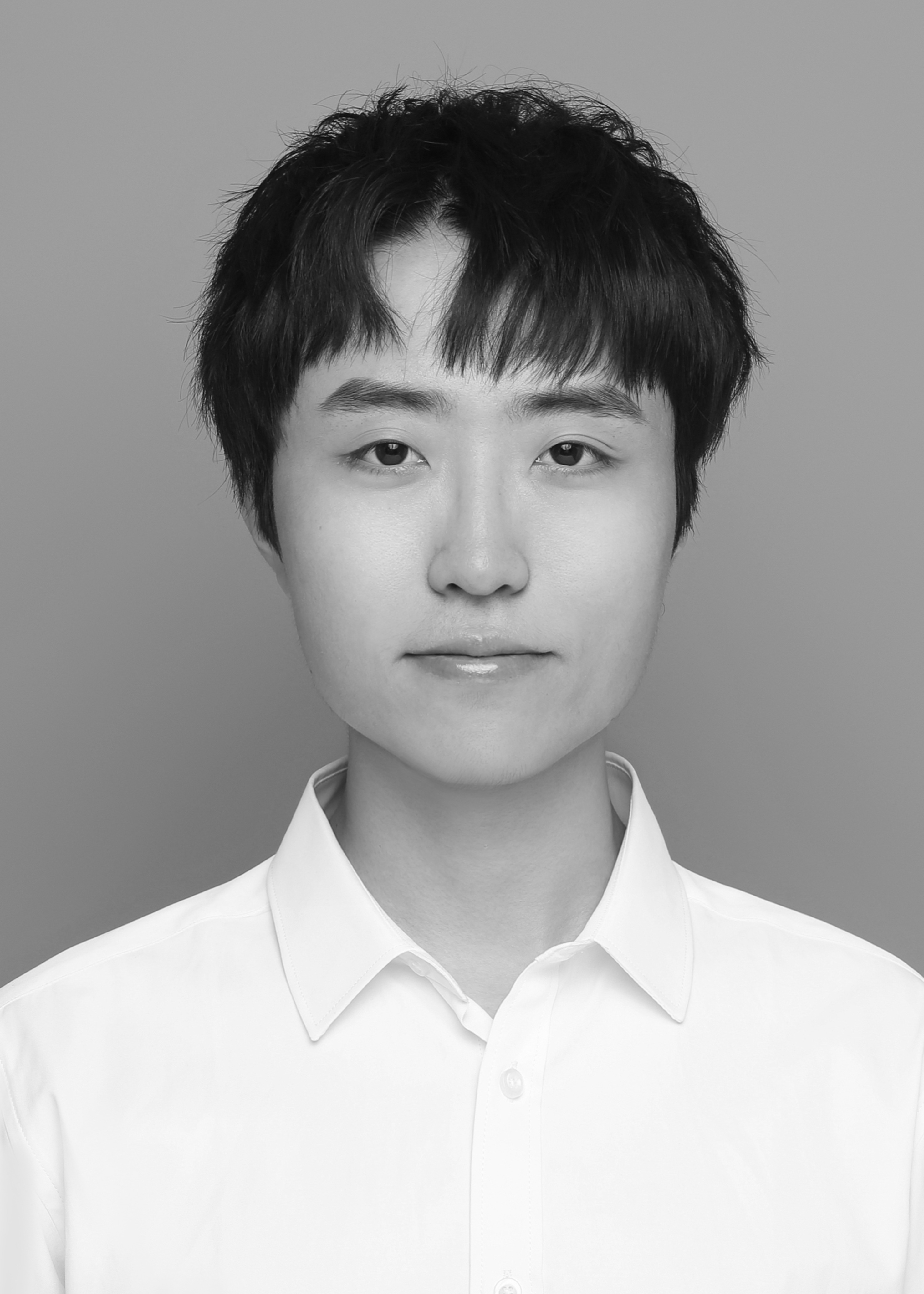}}}]{Yikang He}
received the B.S. degree from the School of Computer Science \& Technology, Beijing Jiaotong University. He is currently a master student at Beijing Jiaotong University. His research interests include 3D face generation and editing.
\end{IEEEbiography}

\vspace{-25pt}
\begin{IEEEbiography}[{\raisebox{0pt}{\includegraphics[width=1in,height=1.25in,clip]{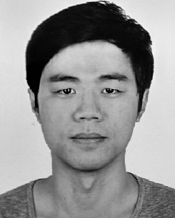}}}]{Jichao Zhang}
is an Assistant Professor with Ocean University of China. He received the Ph.D. degree from Multimedia and Human Understanding Group at the University of Trento. His research interests include deep generative model, 2D and 3D image generation and editing.
\end{IEEEbiography}

\vspace{-25pt}
\begin{IEEEbiography}[{\raisebox{0pt}{\includegraphics[width=1in,height=1.25in,clip]{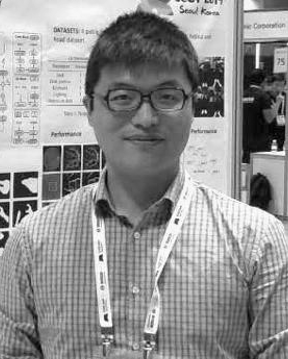}}}]{Wei Wang}
(Member,~IEEE) received the PhD degree from the University of Trento, in 2018. He is a full Professor at the Institute of Information Science at Beijing Jiaotong University. His research interests include machine learning and its application to computer vision and multimedia analysis.
\end{IEEEbiography}

\vspace{-25pt}
\begin{IEEEbiography}[{\raisebox{0pt}{\includegraphics[width=1in,height=1.25in,clip]{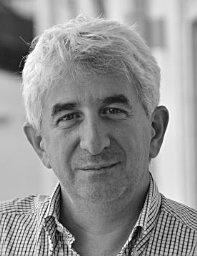}}}]{Nicu Sebe}
(Senior~Member,~IEEE) is Professor with the University of Trento, Italy, leading the Multimedia and Human Understanding Group. He was the General Co-Chair of ACM Multimedia 2013 and 2022, and the Program Chair of  ACM Multimedia 2007 and 2011, ICCV 2017, ECCV 2016 and ICPR 2020. He is a fellow of the International Association for Pattern Recognition.
\end{IEEEbiography}

\vspace{-25pt}
\begin{IEEEbiography}[{\raisebox{0pt}{\includegraphics[width=1in,height=1.25in,clip]{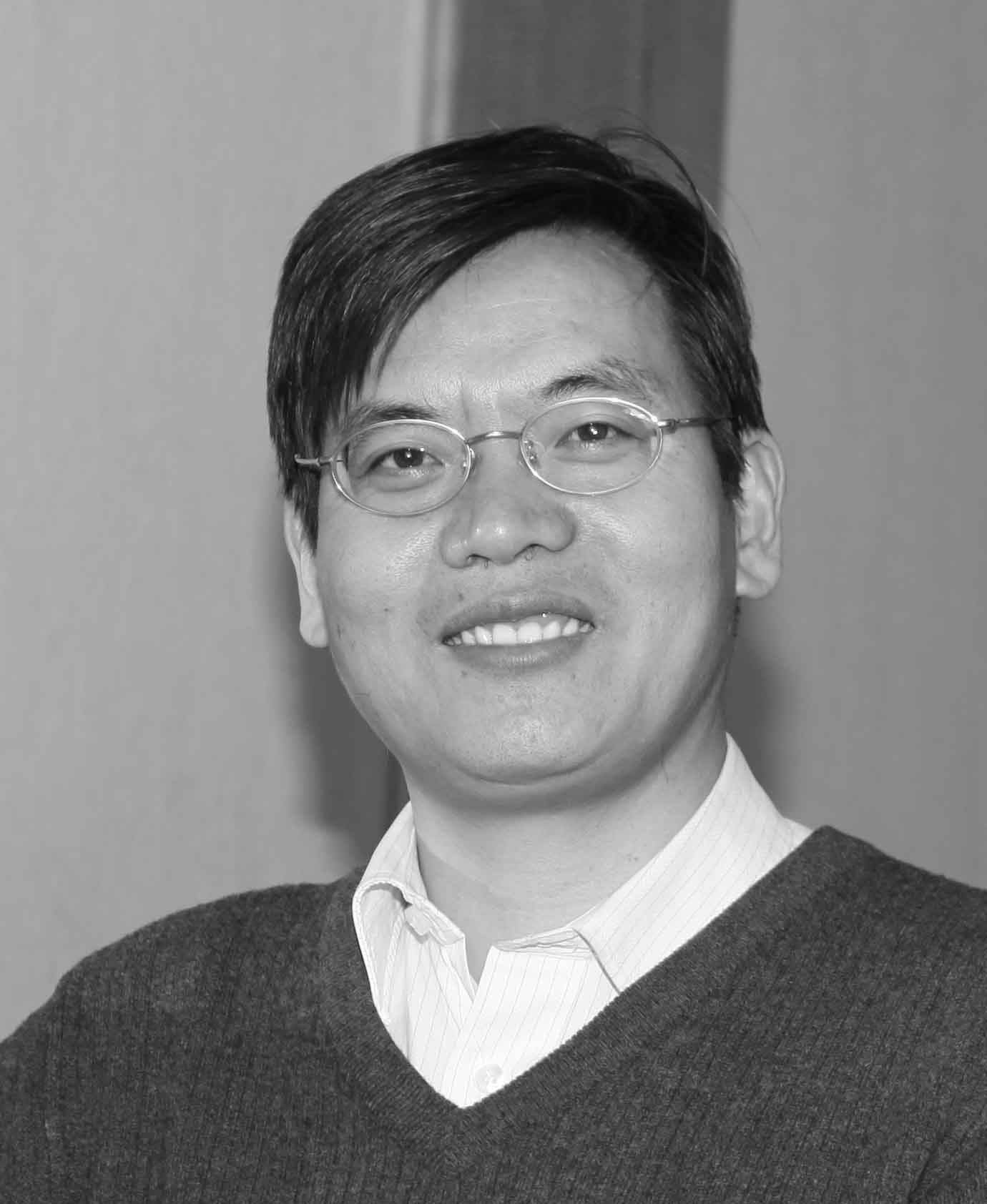}}}]{Yao Zhao}
(Fellow, IEEE) received the B.S. degree from the Radio Engineering Department, Fuzhou University, Fuzhou, China, in 1989, the M.E. degree from the Radio Engineering Department, Southeast University, Nanjing, China, in 1992, and the Ph.D. degree from the Institute of Information Science, Beijing Jiaotong University (BJTU), Beijing, China, in 1996. He is currently the Director of the Institute of Information Science, Beijing Jiaotong University. His current research interests include image/video coding and video analysis and understanding. He was named a Distinguished Young Scholar by the National Science Foundation of China in 2010 and was elected as a Chang Jiang Scholar of Ministry of Education of China in 2013.
\end{IEEEbiography}
\vfill
\end{document}